\documentclass{article}

\PassOptionsToPackage{numbers, compress}{natbib}

\usepackage[table]{xcolor}

 \usepackage[main, final]{neurips_2026}

\usepackage{graphicx}
\usepackage{wrapfig}
\usepackage[utf8]{inputenc} 
\usepackage[T1]{fontenc}    
\usepackage{hyperref}       
\usepackage{url}            
\usepackage{booktabs}       
\usepackage{amsfonts}       
\usepackage[most]{tcolorbox}
\usepackage{nicefrac}       
\usepackage{microtype}      
\usepackage{xcolor}         
\usepackage{multirow}
\usepackage{enumitem}
\usepackage{algorithm}

\usepackage{algpseudocode}
\usepackage{float}
\title{From Retrieval to Reasoning: Agentic Mechanism Prediction from Cell Painting Profiles}

\author{%
  Jiayuan Chen\textsuperscript{1,2}, Botao Yu\textsuperscript{1}\thanks{Equal contribution authors.}, Tianyu Liu\textsuperscript{3}\footnotemark[1], Thai-Hoang Pham\textsuperscript{1,2}\footnotemark[1], \\
    \textbf{Meng Wu\textsuperscript{4},  Ping Zhang\textsuperscript{1,2}}\thanks{Corresponding author.} \\
  \textsuperscript{1}Department of Computer Science and Engineering, The Ohio State University\\
  \textsuperscript{2}Department of Biomedical Informatics, The Ohio State University \\
  \textsuperscript{3}Department of Biostatistics, Yale University\\
  \textsuperscript{4}College of Pharmacy, The Ohio State University \\
\texttt{\{chen.12930,zhang.10631\}@osu.edu} \\
}

\begin{document}

\maketitle

\begin{abstract}
Cell Painting is a high-content morphological profiling assay widely used for phenotype-based biological inference, with mechanism of action (MOA) prediction as a central application. Existing approaches largely formulate Cell Painting-based inference as representation matching, assigning predictions from nearby reference perturbations in morphological feature space. However, retrieved neighbors are often noisy and partially misleading evidence due to batch effects, non-specific cytotoxicity, phenotypic convergence, and source-dependent variability. We reformulate Cell Painting-based MOA prediction as a calibrated evidence reasoning problem, where retrieved neighbors are treated as uncertain observations that must be evaluated, compared, and sometimes rejected before supporting a mechanistic conclusion. We propose PhenoAIR, a reliability-aware multi-agent framework that maintains a candidate-centric evidence memory and performs controller-guided refinement over phenotype- and mechanism-side evidence. PhenoAIR uses offline reference-set calibration to weight evidence by source reliability, phenotype stability, and mechanism-level confusion. We evaluate PhenoAIR on a benchmark\footnote{\url{https://huggingface.co/datasets/Jerrychen229/PhenoAIR_MOA}} constructed from JUMP Cell Painting profiles and annotations, covering controlled, realistic, and discovery-oriented open-world MOA prediction settings. PhenoAIR\footnote{\url{https://github.com/The-Real-JerryChen/PhenoAIR_NeurIPS26}} outperforms representation-matching and LLM-based baselines across all settings.
\end{abstract}
\section{Introduction}

Cell Painting is a high-content screening assay that captures drug- or gene-induced changes in cellular morphology~\cite{bray2016cell,chandrasekaran2024three}, providing a scalable phenotypic readout for downstream biological inference tasks ranging from toxicity assessment~\cite{ewald2026cell} and lead prioritization to functional annotation of genetic perturbations~\cite{chandrasekaran2025morphological,kraus2025rxrx3}. Across these applications, a common computational strategy is to encode perturbations into a morphological representation space and infer biological relationships from the geometry of that space, through clustering, retrieval, label transfer, or prototype matching. In this work, we focus on mechanism of action (MOA) prediction~\cite{caie2010high,ljosa2012annotated}, where existing methods commonly assign labels from the most similar reference perturbations, e.g., via top-$K$ retrieval or distance-weighted voting. This assumes that proximity in phenotype space provides a reliable basis for biological inference. However, similarity alone does not indicate whether the retrieved evidence is biologically informative, reliable, or mechanistically consistent~\cite{chandrasekaran2024three,ramezani2025genome}.

In Cell Painting, morphological similarity is an indirect signal of shared mechanism rather than a mechanistic measurement itself. Similar profiles may reflect a common biological cause, but they may also arise from confounding experimental factors or convergent downstream cellular states. Thus, the core challenge is not retrieving morphologically similar perturbations, but deciding which similarities constitute reliable mechanistic evidence.  This uncertainty creates three requirements that fixed retrieval pipelines do not satisfy. Retrieval quality is instance-dependent: for some queries, relevant mechanisms appear among top-ranked neighbors, whereas for others they are buried deep in the retrieval list, making any fixed retrieval depth inadequate~\cite{arevalo2024evaluating}. Morphological profiles also show source-dependent variability, the same compound profiled across laboratories may yield inconsistent similarity patterns, and cross-source agreement or disagreement is itself an important reliability signal that single-profile retrieval cannot exploit~\cite{chandrasekaran2023jump,seal2025cell}. Finally, the MOA label space is mechanistically structured: polypharmacology and systematically confusable MOA pairs often require pharmacological context to distinguish mechanisms that appear similar in morphology. We therefore argue that \textbf{Cell Painting-based MOA prediction is better formulated not as a matching problem, but as an evidence-based reasoning problem.} Under this view, retrieved neighbors are uncertain observations that must be evaluated, compared, calibrated, and sometimes rejected before they can support a mechanistic conclusion. Importantly, our goal is not to replace Cell Painting with non-phenotypic annotations, but to make phenotypic evidence an explicit object of reasoning: final mechanisms should be accepted only when they are supported by reliable Cell Painting-derived evidence.

Recent LLM and agentic systems offer tool use~\cite{schick2023toolformer}, retrieval-augmented reasoning, and multi-agent collaboration~\cite{du2023improving}, but they are not directly designed for this setting. Biomedical agents~\cite{li2025biomedrag,huang2025biomni} typically use tools for knowledge-centric queries or workflow execution, and generic RAG treats retrieved items as context to summarize. Debate-style multi-agent systems add multiple perspectives, but often rely on free-form consensus rather than persistent candidate-level evidence states. Cell Painting-centered MOA prediction instead requires reasoning over retrieved neighbors whose reliability is uncertain, motivating a task-specific agentic formulation that calibrates evidence, assigns it to explicit hypotheses, and refines decisions through controlled evidence-seeking actions.

To instantiate this evidence-based formulation, we propose PhenoAIR (\textbf{Pheno}typic \textbf{A}gentic \textbf{I}nference with \textbf{R}eliability-aware reasoning), a multi-agent framework for calibrated evidence reasoning over Cell Painting profiles. PhenoAIR does not treat retrieved neighbors as predictions. Instead, it treats them as uncertain evidence for candidate mechanisms and maintains a persistent candidate memory that records which hypotheses are supported, contradicted, suppressed, or still require validation across reasoning rounds. The inference process is organized around this memory. A phenotype-side reasoner interprets calibrated Cell Painting retrieval evidence, while a mechanism-side reasoner independently evaluates the structural and pharmacological plausibility of candidate mechanisms. Their outputs update the shared candidate memory rather than directly determining the final answer. An offline reliability layer further calibrates the evidence by summarizing global reference-set reliability, phenotype stability, and mechanism-level confusion patterns. A rule-based controller then uses the memory state to decide whether to stop, expand retrieval, inspect specific candidates, validate candidate groups, or request additional mechanism-side reasoning. Finally, an arbiter produces the prediction from the full evidence trajectory under explicit eligibility constraints. For open-set MOA inference, where the system must distinguish weak support for known mechanisms from evidence of a genuinely novel mechanism, we extend the controller and arbiter with an offline self-evolving procedure that refines decision rules from observed reasoning trajectories.

To evaluate our approach, we construct a benchmark from the public JUMP Cell Painting Consortium~\cite{chandrasekaran2023jump} and Drug Repurposing Hub~\cite{corsello2017drug}, comprising three complementary settings. MoA-Verified provides a controlled setting with high-confidence cross-source profiles. MoA-Extended reflects realistic deployment conditions with heterogeneous and noisy data. MoA-Novel targets a discovery-oriented setting relevant to first-in-class drug discovery, where a query compound may act through a mechanism not represented in the reference set and methods must recognize when known MOA labels are insufficiently supported. Together, these settings evaluate not only classification accuracy, but also robustness to noisy retrieval evidence and the ability to identify unsupported known mechanisms. Across these settings, PhenoAIR outperforms representation-matching baselines and LLM-based agent/workflow baselines.

Our contributions are as follows:
\begin{itemize}[leftmargin=*]
\item We reformulate Cell Painting-based MOA prediction from representation matching to calibrated evidence reasoning, where retrieved phenotypic
    neighbors are treated as uncertain observations rather than direct answers.

    \item We propose PhenoAIR, a reliability-aware multi-agent framework that replaces free-form agent consensus with candidate-centric evidence memory, externalized reliability calibration, and controller-guided refinement over auditable evidence states.

    \item We construct a curated three-setting benchmark that evaluates MOA prediction under controlled, realistic, and discovery-oriented open-world conditions, and show that PhenoAIR improves over representation-matching and LLM-based baselines.
\end{itemize}

\begin{figure}
    \centering
    \includegraphics[width=0.95\linewidth]{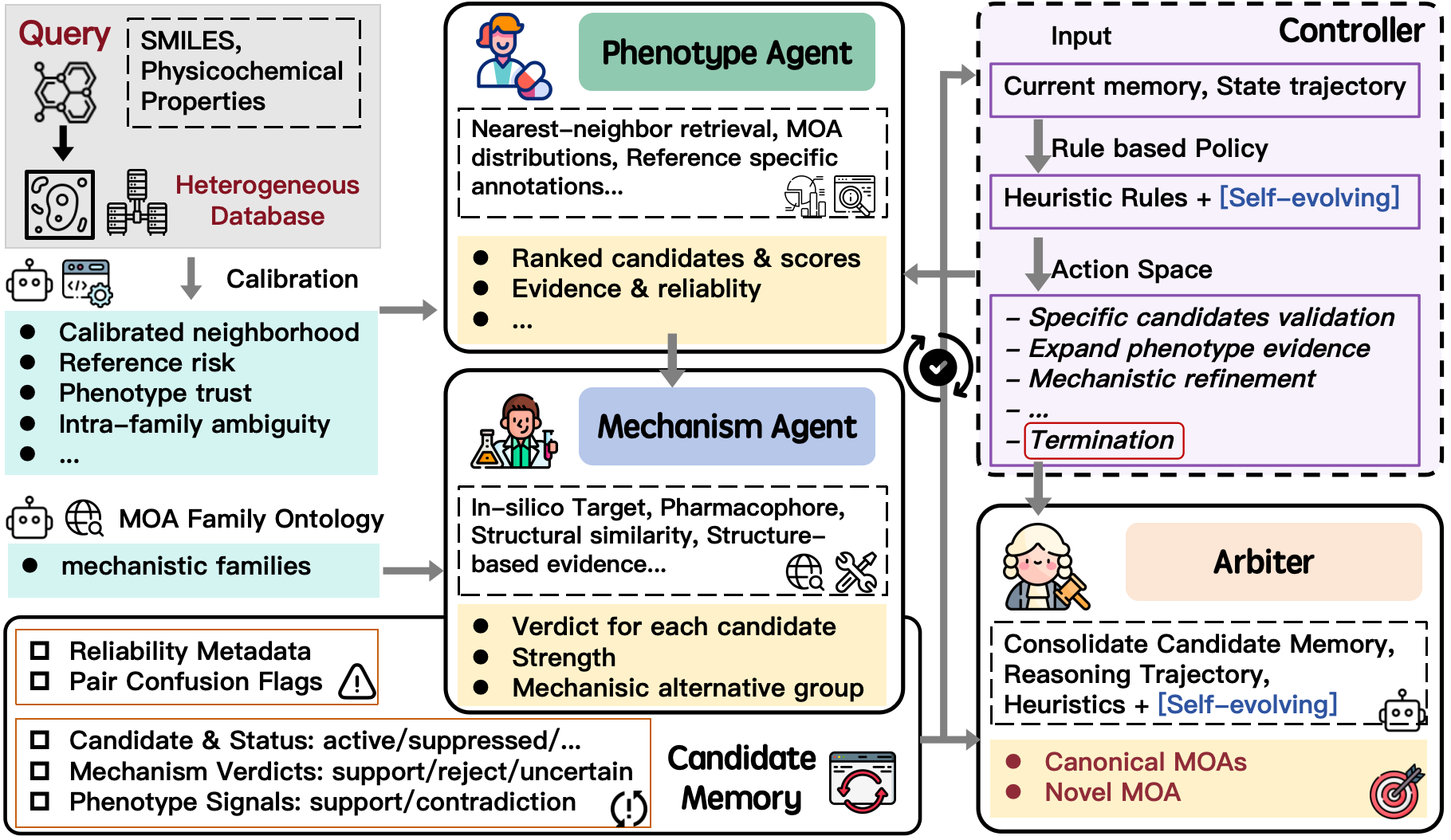}
    \vspace{-6pt}
    \caption{Architecture of PhenoAIR. The system consists of three stages: Externalized reliability calibration (left)  precomputes a reliability-annotated phenotype neighborhood and a mechanistic family ontology from the multi-source reference database.  Candidate-centric evidence reasoning (center) maintains a shared candidate memory updated by Phenotype Agent (calibrated retrieval) and a Mechanism Agent (structural and pharmacological verdicts). Controller-guided refinement (right) iteratively coordinates the two agents through a rule-based controller, terminating with an Arbiter that produces a canonical or novel MOA. Self-evolving extensions (in blue) handle open-set inference. }
    \label{fig:1}
    \vspace{-12pt}
\end{figure}

\section{Method}

\subsection{Problem Formulation}

We consider the task of predicting the mechanism of action (MOA) of a query compound from its Cell Painting morphological profile. Let \(\mathcal{R} = \{(z_i, y_i)\}_{i=1}^N\) denote a reference set, where \(z_i \in \mathbb{R}^d\) is the aggregated morphological profile of reference compound \(i\), and \(y_i \subseteq \mathcal{M}\) is its set of MOA labels drawn from a known label space \(\mathcal{M}\). When a reference compound is profiled across multiple experimental sources, we aggregate its source-specific profiles into a consensus profile \(z_i\) for standard retrieval, while retaining the individual source profiles for cross-source reliability analysis. Given a query compound with Cell Painting profile \(z_q\), a standard retrieval-based formulation obtains a set of nearest reference compounds \(\mathcal{N}_K(q) \subseteq \mathcal{R}\) under a morphological distance function, and predicts \(\hat{y}_q\) by aggregating the MOA labels of the retrieved neighbors, e.g., through top-\(K\) voting or distance-weighted aggregation.

We consider two evaluation settings. In the \textbf{closed-set} setting, the true MOA is assumed to be represented in \(\mathcal{M}\), and the goal is to predict \(\hat{y}_q \subseteq \mathcal{M}\). In the \textbf{open-set} setting, the query may correspond to a mechanism not represented by the known reference label space. The system must therefore either predict a supported known MOA \(\hat{y}_q \subseteq \mathcal{M}\), or output the special label \(\texttt{novel}\) to indicate that the available evidence does not adequately support any candidate in \(\mathcal{M}\).

\subsection{Method Overview}

Recent LLM-based inference systems often combine retrieval, tools, and multi-agent reasoning by treating external evidence as context for generation, discussion, or reflection~\cite{du2023improving,schick2023toolformer,fernando2023promptbreeder}. In Cell Painting-based MOA prediction, however, retrieved evidence is not a set of reliable facts: nearest phenotypic neighbors may reflect batch effects, generic morphology, weak reference anchors, or mechanistically ambiguous profiles. PhenoAIR therefore treats retrieved neighbors as noisy observations whose reliability must be calibrated and whose candidate implications must be tracked.

PhenoAIR replaces label aggregation over nearest neighbors with reliable evidence integration over candidate MOA hypotheses. Given a query profile, the system maintains a candidate memory in which each candidate records phenotype support, phenotype contradiction, mechanism-side verdicts, reliability context, and eligibility state. The framework has three components. First, \textbf{externalized reliability calibration} constructs offline reliability resources and converts raw retrieval into calibrated phenotype evidence. Second, \textbf{candidate-centric evidence reasoning} uses role-separated phenotype and mechanism agents to update shared candidate states rather than conducting free-form debate. Third, \textbf{controller-guided refinement} adaptively selects evidence-collection actions and terminates when the memory state is sufficiently resolved. A final arbiter reasons over the completed memory and trajectory, followed by deterministic eligibility checks that return either a supported canonical MOA or \texttt{novel}. PhenoAIR differs from generic retrieval-augmented generation by treating retrieved neighbors as uncertain observations rather than static context for answer generation. It also differs from debate-style multi-agent systems: rather than relying on unconstrained agent consensus, retrieval, mechanism reasoning, and arbitration are connected through persistent candidate states and deterministic validity checks. Details are provided in Appendices~\ref{app:externalized_reliability}-\ref{app:controller_guided_refinement} and Algorithm~\ref{alg:phenoair_inference}.

\subsection{Externalized Reliability Calibration}
Raw phenotype similarity does not indicate whether a retrieved neighbor is biologically informative. PhenoAIR therefore separates evidence-quality assessment from query-time reasoning through an offline reliability layer. Before inference, it constructs a reference reliability memory \(\mathcal{B}\) from the multi-source reference set. An LLM-assisted code-execution pipeline proposes analyses, computes reference-set measurements, and converts them into structured annotations describing cross-source consistency, reference-anchor replicability, intra-MOA stability, pairwise MOA separation, generic morphology risk, and recurring MOA-confusion patterns. This stage does not observe query outcomes and does not produce query-specific predictions; its outputs are cached before inference.

For a query \(q\), the reliability layer produces a calibrated neighborhood \(\tilde{\mathcal{N}}(q)\) and a query-level reliability context \(r_q\). The calibrated neighborhood serves as ranked phenotype evidence, while \(r_q\) summarizes evidence concentration, source agreement, low-trust anchors, and potential MOA confusions as review-only context. The calibration layer does not introduce new mechanisms or modify the label space.

PhenoAIR also constructs an offline mechanistic family ontology over \(\mathcal{M}\), grouping canonical MOA labels into coarse biological families using external biomedical evidence and label definitions. This ontology supports group-level reasoning when evidence localizes to a family but is insufficient to select a fine-grained canonical label. Final closed-set predictions remain restricted to labels in \(\mathcal{M}\), with \texttt{novel} reserved for open-set cases.

\subsection{Candidate-Centric Evidence Reasoning}
\textbf{Candidate memory.}
Inference maintains $\mathcal{H}_q = \{(c, s_c): c \in \mathcal{C}_q\},$ where \(c\) is a candidate MOA and \(s_c\) is its evidence state. Canonical candidates belong to \(\mathcal{M}\); optional family entries are used only for intermediate validation. Each state stores phenotype support and contradiction, mechanism verdicts (\textit{support}, \textit{reject}, or \textit{uncertain}), candidate status, and reliability metadata. A candidate becomes final-eligible only through accumulated evidence, not merely because it appears in retrieval or is proposed by one agent.

\textbf{Phenotype agent.}
The phenotype agent interprets Cell Painting evidence only. Given \(\tilde{\mathcal{N}}(q)\), \(r_q\), and the current memory, it summarizes local neighbors, broader MOA distributions, and retrieved reference-anchor annotations into candidate-level phenotype evidence. Reference-anchor annotations describe retrieved references, not the query compound, and are used only to assess anchor trustworthiness. The phenotype agent does not use query-side structure, target predictions, or pharmacophore information.

\textbf{Mechanism agent.} The mechanism agent independently evaluates memory candidates using query-side chemical structure, in-silico target predictions, pharmacophore matches, and structural similarity to known compounds. It may propose a canonical candidate or mechanistic family when structure-side evidence supports an alternative, but such proposals are inserted as requiring phenotype grounding or group-level validation rather than becoming directly final-eligible.

\textbf{Memory-mediated communication.}
The agents do not communicate through free-form dialogue. Each reads the current candidate memory and returns structured evidence updates. This makes cross-agent influence explicit: mechanism evidence can challenge a retrieved phenotype hypothesis, phenotype evidence can validate or reject a mechanism-side proposal, and unresolved conflicts remain visible to the controller.

\subsection{Controller-Guided Inference Refinement}

\label{sec:controller_guided_refinement}

Candidate memory records evidence state, but does not decide which evidence should be collected next. PhenoAIR uses a rule-based controller to select auditable refinement actions based on the current memory, reliability context, and trajectory history. The controller does not predict MOA labels; it only chooses the next evidence view. The process runs for query-dependent iterations \(t=0,\ldots,T\), where \(T\) is determined by the controller's STOP action.

\textbf{Iterative refinement and controller policy.}
At iteration \(t\), after phenotype- and mechanism-side updates, the controller selects
\[
a^{(t)}
=
\pi\big(\mathcal{H}_q^{(t)}, r_q, \tau_q^{(t)}\big),
\quad a^{(t)} \in \mathcal{A},
\]
where \(\tau_q^{(t)}\) records past states, agent outputs, and controller actions. The action space contains high-level operations such as expanding phenotype evidence, validating candidates or candidate groups, requesting mechanism-side family reasoning, inspecting risky reference anchors, and terminating inference. The policy uses interpretable signals such as phenotype reliability, phenotype-mechanism agreement, candidate support state, and whether previous actions changed the memory. This avoids both fixed-depth retrieval and free-form LLM planning.

\textbf{Arbiter and final gate.}
When the controller stops, the arbiter produces a candidate-level decision:
\[
y_q^{\mathrm{arb}}
=
\mathrm{Arbiter}\big(\mathcal{H}_q^{(T)}, \tau_q^{(T)}\big).
\]
The arbiter is LLM-based but cannot call evidence-gathering tools. Its output is passed through deterministic eligibility checks:
\[
\hat{y}_q
=
\mathrm{Gate}\big(y_q^{\mathrm{arb}}, \mathcal{H}_q^{(T)}\big).
\]
The gate excludes suppressed candidates, requires adequate support, prevents selection from unsupported calibration metadata alone, and enforces canonical closed-set labels. If no canonical candidate is adequately supported, the output is \(\texttt{novel}\).

\textbf{Self-evolving refinement for open-set inference.} For open-set inference, we use an offline self-evolving procedure over a held-out development set to refine the controller rule registry and the mutable arbiter heuristic section. The procedure does not update LLM parameters, feature extractors, tools, evidence sources, or the label space. Given development trajectories, an offline LLM clusters recurring failure patterns from compressed phenotype, mechanism, controller, memory, and arbiter traces, then proposes schema-constrained controller rules, arbiter heuristic revisions, or retention decisions. Proposed rules must use whitelisted runtime signals and allowed actions; invalid artifacts are blocked. All selected artifacts are fixed before final test evaluation.
\section{Experiment}
\subsection{Benchmark Construction}

We construct a benchmark for Cell Painting-based MOA prediction that under three progressively more challenging regimes: high-confidence closed-set inference, realistic noisy-reference inference, and discovery-oriented open-world inference. The benchmark is built from public Cell Painting profiles and curated drug MOA annotations, allowing us to test not only predictive accuracy, but also robustness to reference quality, source heterogeneity, and incomplete mechanism coverage.

\subsubsection{Data Curation}
We use the cpg0016-jump dataset from the Cell Painting Gallery \cite{chandrasekaran2023jump}, which provides raw Cell Painting images of cells under chemical perturbations. Although CellProfiler well-level features are available, we use image-derived representations from pretrained encoders as the primary morphological profiles, and include feature-based baselines in our comparisons. To capture realistic experimental variability while keeping the benchmark computationally tractable, we select five sources covering approximately 115k unique compounds before MOA annotation filtering. These sources include both overlapping compounds profiled across laboratories and source-specific compounds, enabling evaluation under cross-source batch effects and incomplete source overlap. One source uses a lower perturbation concentration (0.625~\(\mu\)M) than the standard 10~\(\mu\)M used in the other sources, further introducing dosage-dependent phenotypic variation.

Ground-truth MOA annotations are obtained from the Broad Institute Drug Repurposing Hub \cite{corsello2017drug}. We match compounds to the Hub and retain those with valid annotations. Since compounds may exhibit polypharmacology, we use curated multi-label MOA sets for evaluation. To ensure label quality, we remove redundant or hierarchically dependent annotations so that remaining labels correspond to distinct mechanistic hypotheses (details in Appendix~\ref{app:dataset}). In addition, each compound is assigned a representative MOA label for compatibility with standard retrieval-based baselines, while all primary evaluations also include the full multi-label annotation set.

\subsubsection{Evaluation Settings}
We define three complementary evaluation settings. \textbf{MoA-Verified}
prioritizes reference reliability: the reference set contains compounds
profiled across multiple sources, providing higher-confidence and more
reproducible morphological evidence. This setting evaluates performance under relatively clean retrieval conditions. \textbf{MoA-Extended} prioritizes MOA coverage: it includes all available annotated compounds regardless of source count, expanding the label space while introducing heterogeneous reference quality, including single-source profiles that are more susceptible to batch effects. This setting evaluates robustness under realistic noisy retrieval conditions. In both settings, we retain MOA classes with at least two distinct reference compounds to reduce source-specific artifacts and support meaningful similarity-based evaluation. \textbf{MoA-Novel} evaluates discovery-oriented open-world inference. In this setting, a subset of test compounds (50\%) is associated with mechanisms not represented in the reference label space. Models must therefore predict known MOAs for in-distribution compounds while recognizing cases where available reference MOAs are insufficiently supported, instead of forcing a nearest-neighbor label. The test set contains a balanced mixture of known-MOA and novel-MOA cases. Dataset statistics are reported in \autoref{tab:1}.

We use different metrics for closed-set and open-world settings. For MoA-Verified and MoA-Extended, we report \textbf{Top-1 accuracy} with respect to the representative MOA label of each compound, and \textbf{Any-match accuracy}, where a prediction is correct if it matches any label in the curated multi-label ground truth. For MoA-Novel, we report \textbf{overall accuracy}, treating \texttt{novel} as an additional output class, and \textbf{Novel detection F1}, which evaluates known-versus-novel discrimination as a binary classification task. For representation-based baselines (e.g., kNN), predictions are matched to ground-truth labels by exact string comparison. For LLM-based methods that will produce open-form textual outputs, we first apply string-based matching to map predictions to canonical MOA labels; unmatched cases are resolved using an LLM-as-judge, which we verify to be highly consistent with human expert annotations (details in Appendix~\ref{app:imple}).

\begin{table}[t]
\vspace{-0pt}
    \centering
        \caption{Benchmark statistics. Our benchmark comprises three settings constructed from the JUMP Cell Painting Consortium and Drug Repurposing Hub.}
        \vspace{-3pt}
    \begin{tabular}{l|cccccc}
        \toprule
       Dataset  & \# Reference & \# Test &\# MOA classes & Avg Sources/Ref. & Avg Labels \\
       \midrule
           MoA-Novel & 63 & 100 & 26 (Known) & 4.05 & 1.00 \\
    MoA-Verified     & 63 & 200 & 26 & 4.05 & 1.19\\
    MoA-Extended & 533 & 600 & 137 & 2.64 &1.13\\
    \bottomrule
    \end{tabular}
    \label{tab:1}
    \vspace{-15pt}
\end{table}
\subsubsection{Baselines}
We consider baselines spanning representation-based retrieval and LLM-based inference paradigms.
\textbf{Representation-Based Retrieval.}
We evaluate standard nearest-neighbor retrieval methods in morphological and molecular representation spaces. For Cell Painting, we use (i) CellProfiler features provided by the Broad Institute, and (ii) deep learning representations from foundation models including CellCLIP~\cite{lucellclip} . Predictions are obtained by distance-weighted voting over the top-$K$ neighbors, following established practice in phenotypic profiling~\cite{caie2010high}. 
In addition, we include a structure-based retrieval baseline using molecular fingerprints to identify structurally similar compounds and transfer their MOA labels.

\textbf{LLM-based inference.}
We evaluate LLM-based baselines that use chemical and phenotypic evidence. The \textbf{structure-only} baseline uses structure-driven information and external mechanistic signals without access to phenotypic profiles. It serves as a diagnostic control for non-phenotypic MOA signal, rather than as a Cell Painting analysis method. We further include a fixed \textbf{workflow} baseline that follows a predefined phenotype-then-mechanism reasoning sequence. The LLM first analyzes Cell Painting retrieval evidence and then incorporates structure-driven information to make a final prediction. Unlike PhenoAIR, this workflow does not maintain a persistent candidate memory, adaptively expand or validate evidence, or revise its reasoning through controller-guided refinement. Finally, we include a \textbf{single-agent} variant that has access to the same evidence tools as PhenoAIR but performs inference within a single agent. This baseline isolates the effect of reasoning decomposition, shared candidate memory, and controller-guided refinement.

\subsubsection{Implementation Details}
For representation-based retrieval, we use well-level Cell Painting features with post-processing to mitigate batch effects, followed by aggregation into compound-level representations across sources. We evaluate multiple aggregation strategies and report the best-performing variant. The kNN baselines use cosine distance for Cell Painting features and Tanimoto distance on Morgan fingerprints (radius 2, 2048 bits) for the ECFP baseline, and predict by inverse-distance-weighted voting over the $K=5$ nearest reference compounds. For LLM-based methods, we evaluate both GPT-5.1 and Claude 4 Sonnet. We choose them because of their stable tool-use capability, reasoning ability, and biological knowledge, while reducing reliance on memorized associations between chemical structures and MOA labels. The goal is to evaluate inference from external evidence rather than recall of known structure-function mappings. The structure-only baseline has access to structure-driven and mechanism-side tools without phenotype information, while the fixed-workflow baseline uses the same evidence tools as PhenoAIR but follows a predefined inference pipeline. To reduce information leakage, we curate external knowledge sources and exclude trivial tool outputs that directly reveal ground-truth labels, such as exact target-gene matches with confidence 1.0. All
LLM-based methods use consistent tool interfaces and comparable prompt settings. For stochastic LLM-based experiments, we report results averaged over five independent runs. Additional details on prompts, tool definitions, post-processing steps, computational cost, runtime, and API usage are provided in the Appendix~\ref{app:cost}.

\subsection{Main Results}
\label{sec:main-results}
\begin{table}[t]
    \centering
        \caption{Main results on three datasets. We report 
accuracy (Acc. \%), hit rate (Any-match), and F1 measured against ground-truth MOA labels. All LLM-based methods 
operate on the same CellCLIP features. 
\textbf{Best} results are in bold; \underline{second-best} are 
underlined. Our method is highlighted in blue.}
\resizebox{1.0\textwidth}{!}{
    \begin{tabular}{c|c|cc|cc|cc}
    \toprule
   &   Dataset   & \multicolumn{2}{c|}{MoA-Verified}  &   \multicolumn{2}{c|}{MoA-Extended} & \multicolumn{2}{c}{MoA-Novel} \\
   &   Metrics   & Acc & Any-match & Acc & Any-match &  Acc & F1 (Novel)\\
      \midrule
\multirow{4}{*}{\rotatebox{90}{kNN} }    &
Random & $3.5\pm1.4$ & $12.3\pm 2.0$ &$1.2\pm0.3$ & $3.2\pm0.5$ & \multicolumn{2}{c}{\multirow{4}{*}{{N/A} }} \\

& CellProfiler    & $9.5\pm0.0$ & $23.5\pm0.0$ & $6.8\pm0.0$ & $12.7\pm0.0$\\
     &CellCLIP  & $14.0\pm0.0$ & $26.5\pm0.0$ & $10.2\pm0.0$ & $16.0\pm0.0$\\
      &ECFP & $23.0\pm0.0$ & $41.0\pm0.0$ & $30.0\pm0.0$ & $42.3\pm0.0$\\    
     \midrule
\multirow{4}{*}{\rotatebox{90}{GPT 5.1}}    & Structure-only   & $43.5\pm1.4$ & $59.0\pm1.1$ & $39.8\pm0.3$ & $52.9\pm0.1$ & $44.2\pm3.4$ & $53.3\pm4.0$\\
 & Workflow & $50.0\pm1.2$ & $56.0\pm1.0$ & $31.7\pm0.5$ & $35.4\pm0.3$ & $58.0\pm0.3$ & $61.3\pm0.8$\\
 & Single Agent & $39.1\pm1.4$ & $59.5\pm1.1$ & $30.9\pm0.5$ & $46.2\pm0.5$ & $48.8\pm2.4$ & $67.3\pm1.1$\\
 & PhenoAIR \cellcolor{blue!20} & $\mathbf{62.6\pm0.7}$ \cellcolor{blue!20} & $\mathbf{74.6\pm0.6}$ \cellcolor{blue!20} & $\underline{45.6\pm0.2}$ \cellcolor{blue!20} & $\underline{56.2\pm0.3}$ \cellcolor{blue!20} & $\mathbf{79.4\pm2.2}$ \cellcolor{blue!20} & $\mathbf{74.7\pm2.9}$ \cellcolor{blue!20}\\
 \midrule
\multirow{4}{*}{\rotatebox{90}{Claude 4} }  & Structure-only & $47.3\pm0.8$ & $64.5\pm1.0$ & $40.8\pm0.3$ & $55.7\pm0.2$ & $43.2\pm1.1$ & $50.6\pm1.0$\\
 & Workflow & $52.4\pm0.9$ & $60.0\pm0.7$ & $37.7\pm0.6$ & $42.2\pm0.3$ & $63.6\pm1.5$ & $70.2\pm1.2$\\
  & Single Agent & $40.5\pm 1.4 $& $58.3\pm0.9$& $32.1\pm0.7$& $45.5\pm0.5$&$52.0\pm1.2$ & $66.9\pm1.1$\\
 & PhenoAIR \cellcolor{blue!20}& \underline{$61.5\pm0.5$} \cellcolor{blue!20} & $\underline{74.2\pm0.7}$ \cellcolor{blue!20} & $\mathbf{45.8\pm0.3}$ \cellcolor{blue!20} & $\mathbf{59.9\pm0.1}$ \cellcolor{blue!20} & \cellcolor{blue!20} $\underline{73.6\pm3.1}$ & \cellcolor{blue!20} $\underline{74.1\pm2.0}$\\
 \bottomrule
    \end{tabular}}
\vspace{-18pt}
    \label{tab:2}
\end{table}

\autoref{tab:2} reports results on the three evaluation settings. LLM-based methods are averaged over five independent runs; deterministic retrieval baselines are reported without variance. Overall, representation-based retrieval methods perform substantially worse than LLM-based inference methods. Among kNN baselines, learned Cell Painting representations (CellCLIP) improve over CellProfiler features, while ECFP-based chemical retrieval is strongest within this group. However, all retrieval-only methods remain limited, indicating that neither morphological nor structural nearest-neighbor transfer is sufficient for reliable MOA prediction.

LLM-based baselines show that simply adding more evidence is not enough.
Structure-only inference provides a strong baseline, reflecting the predictive value of chemical and mechanistic information. However, the fixed workflow that combines phenotype retrieval with structure-side evidence does not consistently improve over structure-only inference, especially on MoA-Extended, where retrieval evidence is noisier. This suggests that uncalibrated phenotype retrieval can introduce misleading evidence when incorporated through a fixed pipeline. The single-agent baseline also underperforms PhenoAIR, showing that the gains come not merely from tool access, but from role-separated multi-agent reasoning
coupled with structured candidate memory and controller-guided refinement.

PhenoAIR achieves the best performance across all three settings and both LLM backbones. On MoA-Verified and MoA-Extended, it improves over both representation-matching baselines and LLM-based baselines, demonstrating the benefit of calibrated evidence reasoning for closed-set MOA prediction. The advantage is most pronounced in MoA-Novel: while retrieval baselines are not  applicable because they must assign each query to an existing reference label, PhenoAIR substantially improves both overall accuracy and novel-detection F1. These results indicate that candidate-centric memory and reliability-aware arbitration help the model recognize when known MOA labels are insufficiently supported, rather than forcing a prediction from noisy or incomplete evidence.

\subsection{Ablation Studies}
\subsubsection{Component Ablation}
\begin{wraptable}{h}{0.5\linewidth}
\vspace{-38pt}
    \centering
        \caption{Effect of core components on two datasets. Variants are constructed by incrementally adding 
each PhenoAIR component to a base LLM, isolating the contribution of 
reliability calibration, candidate-centric memory with multi-agent 
reasoning. All variants use the 
same base LLM and CellCLIP features. Full results are in Appendix.}
    \begin{tabular}{c|lcc}
    \toprule
       &   Dataset   &Verified  &   {Extended}  \\
   &   Method   & Acc  & Acc  \\
    
    \midrule
  \multirow{4}{*}{\rotatebox{90}{GPT 5.1}}    & Base LLM & $16.4$ & $18.7$\\
& PhenoAIR-Base &$55.6$ & $43.9$\\
         & + Memo. & $60.1$ & $44.5$ \\
  & + Memo. \& Cal. & $62.6$ & $45.6$\\
         \midrule
  \multirow{4}{*}{\rotatebox{90}{Claude 4} }     
           & Base LLM & $26.7$ & $23.3$ \\
         & PhenoAIR-Base & $55.0$ & $42.3$\\
         & + Memo. & $59.3$ & $44.2$ \\
    & + Memo. \& Cal.& $61.5$ & $45.8$ \\
         \bottomrule
    \end{tabular}
    \vspace{-6pt}
    \label{tab:ablation}
\end{wraptable}
\autoref{tab:ablation} reports an ablation that incrementally adds PhenoAIR components on top of a base LLM. Direct prompting with only SMILES information performs poorly on both benchmarks, indicating that parametric knowledge alone is insufficient for fine-grained MOA prediction. Adding the base PhenoAIR multi-agent workflow, without candidate memory or reliability calibration, substantially improves performance, showing that structured phenotype- and mechanism-side evidence reasoning is already much stronger than direct prompting.

Candidate-centric memory provides consistent additional gains across both LLM backbones and datasets. This suggests that explicitly tracking candidate hypotheses and their supporting or contradicting evidence across rounds is beneficial beyond within-turn agent context. Adding reliability calibration further improves performance, especially on MoA-Verified, where the calibrated reference evidence is more stable and reliable. The smaller gain on MoA-Extended is consistent with its harder retrieval regime: many queries have weak or deeply ranked phenotype evidence, limiting what can be recovered from local retrieval even with calibrated reasoning.

\begin{figure}[t]
    \centering
\vspace{-6pt}
    \includegraphics[width=0.35\linewidth]{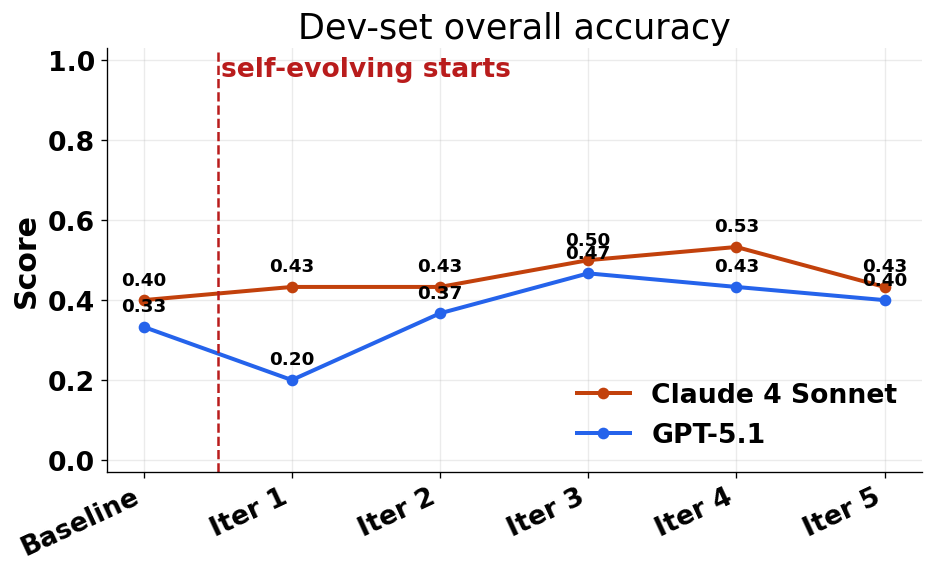}
    \hfill
    \includegraphics[width=0.35\linewidth]{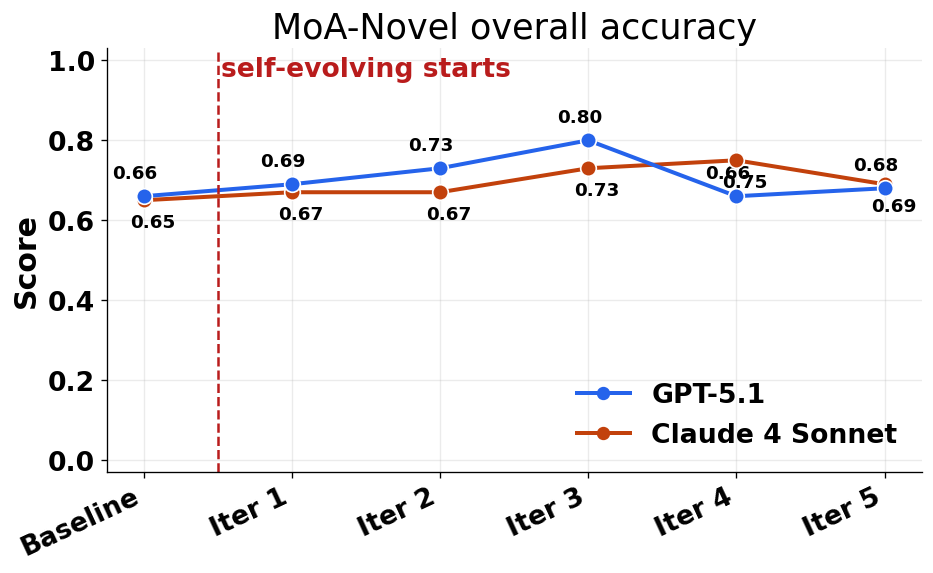}
    \hfill
    \includegraphics[width=0.26\linewidth]{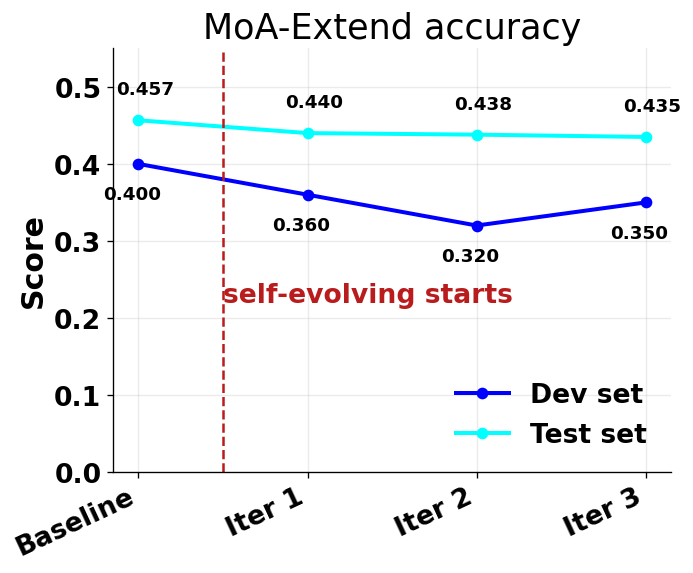}
\vspace{-6pt}
    \caption{
Self-evolving refinement curves across iterations. Controller rules and arbiter prompts are updated using development-set trajectories after the baseline step. Left two: development set and test set overall accuracy on MoA-Novel. Right: MoA-Extended accuracy on development and test sets.
    }
    \label{fig:2}
    \vspace{-7pt}
\end{figure}
\subsubsection{Effect of Self-Evolving Refinement}
\autoref{fig:2} evaluates self-evolving refinement, where controller rules and arbiter prompts are iteratively updated using trajectories from a small development set. On MoA-Novel, refinement improves both GPT-5.1 and Claude 4 Sonnet, suggesting that trajectory-driven updates help distinguish supported known mechanisms from novel cases. The gains are not monotonic, indicating that later iterations can overfit to development trajectories rather than uniformly improve inference. On MoA-Extended, refinement does not improve accuracy and sometimes degrades performance, likely because its failure modes are broader and noisier, making rules from a small development set less transferable. We therefore restrict self-evolving refinement to the open-set MoA-Novel setting, where it targets the additional novel-detection decision, and do not apply it to the closed-set MoA-Verified or MoA-Extended settings to avoid dev-set-specific rule tuning for standard classification.
\section{Analysis and Discussion}
\subsection{Confidence Calibration}
To test whether PhenoAIR's gains rely on Cell Painting evidence rather than structure-side information, we shuffle one evidence stream across compounds while keeping the rest of the pipeline unchanged. Using GPT-5.1, we evaluate 100 MoA-Extended queries whose ground-truth MOA ranks highest in the calibrated phenotype neighborhood, i.e., where phenotype evidence is informative.
\begin{table}[h]
\centering
\small
\caption{Evidence-shuffling control on MoA-Extended (GPT-5.1, mean $\pm$ s.d.
over five runs).}
\label{tab:shuffle}
\begin{tabular}{lccc}
\toprule
Setting & Phenotype & Structure side & Acc. (\%) \\
\midrule
Full PhenoAIR           & matched  & matched  & 80.0 $\pm$ 1.8 \\
Shuffled structure side & matched  & shuffled & 63.2 $\pm$ 0.7 \\
Shuffled phenotype      & shuffled & matched  & 53.6 $\pm$ 1.0 \\
Structure-only agent    & removed  & matched  & 50.2 $\pm$ 0.7 \\
\bottomrule
\end{tabular}
\end{table}

Shuffling only the phenotype evidence reduces accuracy by 26.4 points, to a level close to the structure-only agent, even though the structure-side inputs and reasoning pipeline are unchanged. This shows that PhenoAIR's gains depend on correctly matched, compound-specific phenotype evidence rather than on chemical structure alone. Conversely, PhenoAIR retains 63.2\% accuracy when structure-side evidence is shuffled, indicating that the two evidence streams are complementary.

\subsection{Robustness Across Image Features}
\begin{figure}[t]
\vspace{-4pt}
    \centering
    \includegraphics[width=0.5\linewidth]{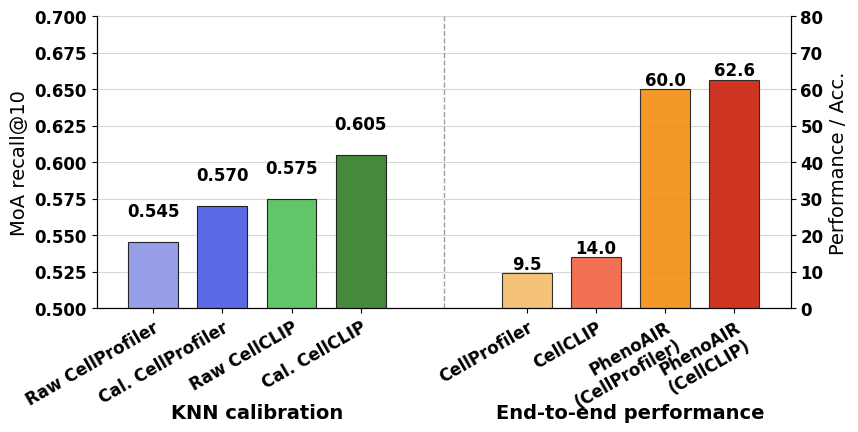}
    \hfill
    \includegraphics[width=0.45\linewidth]{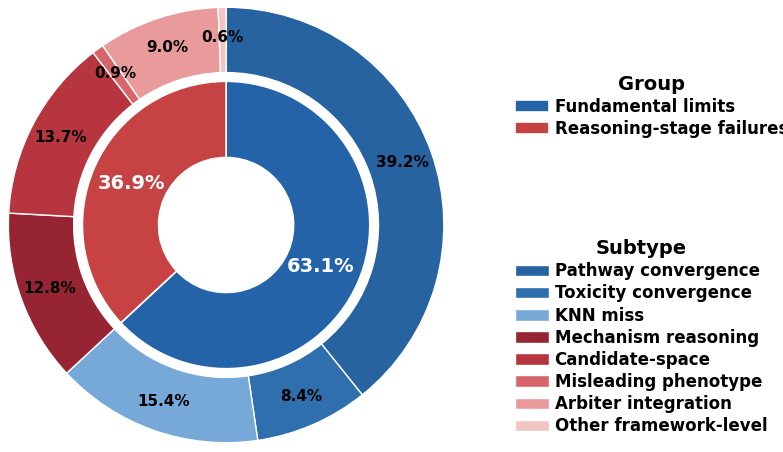}
    \hfill
        \vspace{-6pt}
    \caption{
{(Left)} MoA Recall@10 with and without calibration, and  accuracy comparison on MoA-Verified, both shown across CellProfiler 
and CellCLIP features. 
{(Right)} Failure mode distribution on MoA-Extended. Inner ring: high-level categories; outer ring: subtype breakdown.}
    \label{fig:robustness-failure}
    \vspace{-10pt}
\end{figure}

We test whether PhenoAIR depends on a particular morphological feature space by comparing CellProfiler features with CellCLIP embeddings (\autoref{fig:robustness-failure}). Calibration yields only modest retrieval improvements, increasing Recall@10 from 0.545 to 0.570 for CellProfiler and from 0.575 to 0.605 for CellCLIP. This indicates that calibration is not mainly a re-ranking mechanism. The small change in Recall@10 also suggests that the downstream gains are not explained by improved neighbor ranking alone; they likely depend on the additional reliability context attached to the retrieved evidence. End-to-end, PhenoAIR reaches 60.0\% accuracy with CellProfiler features and 62.6\% with CellCLIP, compared with 9.5\% and 14.0\% for kNN on the same features. The gap between feature spaces is small relative to either configuration's margin over kNN, indicating that PhenoAIR's performance is not tied to a particular morphological representation.

\subsection{Limitations and Failure Modes}
We categorize failure cases on MoA-Extended in \autoref{fig:robustness-failure}. Most errors arise from fundamental evidence limitations, including pathway-level convergence, toxicity-driven convergence, and retrieval misses, where the local phenotype neighborhood either reflects non-specific cellular responses or lacks sufficient evidence for the correct MOA. This supports our central motivation: morphological similarity is not a mechanistic explanation, and Cell Painting-based MOA prediction requires explicit reasoning about evidence reliability. The remaining errors come from reasoning-stage failures, such as mechanism misinterpretation, candidate-space errors, misleading phenotype evidence, and imperfect arbiter integration. These failures indicate that PhenoAIR is not a complete solution to mechanistic interpretation, but they also identify actionable directions for future work, including stronger mechanism grounding, candidate validation, and arbitration under conflicting evidence.

\section{Conclusion}
In this paper, we argue that Cell Painting-based MOA prediction should be treated as an evidence reasoning problem rather than a direct representation-matching task. PhenoAIR instantiates this view by evaluating the reliability and consistency of retrieved phenotypic evidence before committing to a mechanistic prediction. Across controlled, realistic, and open-world settings, our results show that structured reasoning is most useful when similarity-based evidence is noisy, incomplete, or ambiguous. These findings suggest that future phenotype-based discovery systems should complement improved representations with explicit mechanisms for evidence evaluation and decision-making. 
\section*{Acknowledgements}
This work was funded in part by the National Science Foundation under award number IIS-2145625 and by the National Institutes of Health under award numbers R01AI188576 and R21EB030294.
{
\bibliographystyle{apalike}
    \bibliography{main}
}

\newpage
\appendix

\section*{Appendix}

\addcontentsline{toc}{section}{Appendix}

Section~\ref{app:related} reviews prior work on Cell Painting morphological profiling and the use  of large language models for biological reasoning, contextualizing PhenoAIR within both lines of research.

Section~\ref{app:results} provides additional quantitative results and qualitative case analyses that complement the main paper.

Section~\ref{app:framework} provides algorithmic details of each PhenoAIR component, including the construction of the externalized reliability layer, the candidate-centric memory schema, the controller's rule set, the arbiter's eligibility constraints, and the self-evolving refinement procedure. 

Section~\ref{app:benchmark} describes how the benchmark was constructed from raw data, how ground-truth MOA labels were curated, and how predictions are evaluated across the three settings.

Section~\ref{app:prompts} lists the full prompts used for each agent (Phenotype Agent, Mechanism Agent, Arbiter, and the offline calibration LLM), along with the schema definitions for all tools available to the agents during inference.

\section*{Algorithm Summary}
Algorithm~\ref{alg:phenoair_inference} summarizes the full inference procedure. We separate global pre-computation from query-time inference. The reliability memory, calibrated retrieval rule, and mechanistic family ontology are constructed once using the reference set and development split. At test time, a query only invokes the fixed calibrated retrieval module, the phenotype and mechanism agents, the controller, and the final arbiter. For open-set experiments, the controller rule registry and arbiter heuristic section may be selected by the offline self-evolving procedure described in Appendix~\ref{app:self_evolving_procedure}. These artifacts are fixed before final test evaluation and are not updated using test outcomes.

\section{Related Works}
\label{app:related}
\subsection{Cell Painting Profiling}
High-content imaging assays such as Cell Painting enable large-scale characterization of cellular responses to chemical and genetic perturbations through multiplexed microscopy measurements of cellular morphology~\cite{bray2016cell,chandrasekaran2024three,chandrasekaran2023jump,kraus2025rxrx3}. Early work demonstrated that high-dimensional morphological profiles could capture biologically meaningful drug response signatures and support mechanism-of-action (MOA) analysis through clustering and nearest-neighbor similarity analysis~\cite{caie2010high}.  

Recent progress has focused heavily on representation learning for cellular morphology. Traditional approaches relied on hand-crafted features~\cite{carpenter2006cellprofiler} extracted from cellular segmentation and intensity statistics, while more recent methods leverage vision foundation models and self-supervised learning to learn scalable microscopy representations. In particular, large-scale self-supervised microscopy foundation models based on masked autoencoding and Vision Transformers have substantially improved biological signal recovery and cross-dataset generalization~\cite{kraus2024masked,chen2025integrating}.   Contrastive multimodal learning has further extended this paradigm by aligning cellular phenotypes with molecular structure representations in a shared latent space~\cite{sanchez2023cloome}. Recent works such as MolPhenix~\cite{fradkin2024molecules} and CellCLIP~\cite{lucellclip} formulate cross-modal phenomolecular retrieval as a contrastive learning problem, enabling zero-shot retrieval of molecular structures from microscopy phenotypes.  These learned representations have enabled a broad range of downstream applications, including MOA prediction, perturbation similarity analysis, target and pathway inference, virtual screening, and cross-modal retrieval between molecules and phenotypic profiles. However, despite substantial advances in representation quality, downstream inference in most existing pipelines still fundamentally relies on representation matching. Treatments are typically compared through nearest-neighbor retrieval, similarity ranking, or embedding-space aggregation, where biological conclusions are derived from phenotypic proximity in the learned representation space~\cite{moshkov2024learning}.

Recent work has increasingly recognized that this paradigm is heavily affected by confounding technical and biological variation. Multiple studies show that learned representations simultaneously capture both phenotypic signal and nuisance variation such as batch, source, and experimental artifacts, motivating extensive efforts on batch correction, representation disentanglement, and domain generalization.  In practice, realistic phenotypic neighborhoods frequently contain noisy or mechanistically heterogeneous neighbors due to polypharmacology, inactive perturbations, cytotoxicity, source-specific variation, and phenotypic convergence~\cite{ramezani2025genome,fradkin2024molecules,arevalo2024evaluating}. Nevertheless, most existing approaches address these challenges primarily through improved feature learning, retrieval calibration, or batch correction, while continuing to treat retrieved neighbors as direct evidence for prediction~\cite{tang2024morphological}. In contrast, our work argues that Cell Painting-based MOA prediction should be formulated not purely as a representation matching problem, but as an evidence reasoning problem over noisy phenotypic observations. Rather than directly aggregating nearest neighbors in embedding space, our framework treats retrieved phenotypic matches as uncertain and potentially conflicting evidence that must be actively evaluated, filtered, and integrated through agentic inference.
\subsection{LLMs for Biological Reasoning}

LLMs are increasingly being explored as reasoning systems for biological and biomedical research. Early efforts primarily focused on biomedical adaptation through domain-specific pretraining~\cite{cui2024scgpt} and retrieval-augmented generation~\cite{soman2024biomedical}, enabling LLMs to access biological literature, pathway databases, and structured biomedical knowledge. More recent work has shifted toward agentic scientific systems that combine iterative reasoning, tool use, and autonomous workflow execution for complex biological tasks.

Recent systems~\cite{averly2025liddia,yang2025multi} formulate drug discovery as an autonomous reasoning process in which LLM agents iteratively explore chemical space and optimize molecular properties through external computational tools.  The Virtual Lab~\cite{swanson2025virtual} further extends this paradigm toward collaborative AI scientific teams, where multiple agents coordinate across hypothesis generation, protein modeling, and experimental planning to design novel SARS-CoV-2 nanobodies with experimental validation.   Other recent frameworks~\cite{su2025biomaster,huang2025biomni} investigate multi-agent architectures for automating bioinformatics workflows, tool orchestration, error recovery, and computational analysis pipelines.  Collectively, these works suggest a broader transition from static prediction pipelines toward interactive and evidence-driven biological reasoning systems. In parallel, multimodal biological foundation models such as Geneverse~\cite{liu2024geneverse} and related genomic-proteomic LLM systems have explored integrating heterogeneous biological modalities within unified representation and reasoning frameworks. These approaches demonstrate the growing interest in combining large language models with multimodal biological data and structured scientific workflows.

However, despite recent progress in biological agent systems and multimodal biomedical LLMs, Cell Painting pipelines remain largely dominated by representation matching approaches based on nearest-neighbor retrieval, similarity aggregation, and embedding-space ranking over phenotypic representations. Existing work primarily focuses on improving representation quality, retrieval performance, or batch correction, while explicit reasoning over noisy and potentially conflicting phenotypic evidence remains largely unexplored~\cite{seal2025cell,wong2023deep}. In contrast, our work formulates Cell Painting-based MOA prediction as an evidence reasoning problem in which retrieved phenotypic neighbors must be iteratively evaluated, filtered, and integrated through coordinated phenotypic and mechanistic reasoning

\section{Extended Experimental Results}
\label{app:results}
\subsection{Additional Quantitative Results}
\subsubsection{Full Benchmark Results}
Tables~\ref{tab:novel-results} and~\ref{tab:ablation-verified-extended} 
report the full benchmark results across two backbone LLMs (GPT-5.1 and 
Claude~4), complementing the main results presented in 
Section~\ref{sec:main-results}. On MoA-Novel (Table~\ref{tab:novel-results}), 
we evaluate Precision, Recall, and F1 of novel-MOA prediction in 
addition to Overall Accuracy, and include ablations of self-evolving 
refinement and reliability calibration to isolate their respective 
contributions to open-set inference. On MoA-Verified and MoA-Extended 
(Table~\ref{tab:ablation-verified-extended}), we provide a stepwise 
ablation that progressively introduces PhenoAIR's core components:
multi-agent reasoning, candidate evidence memory, and reliability 
calibration onto the Base LLM. Across both tables, PhenoAIR 
consistently improves over its ablated variants and over non-agentic 
baselines on both backbones, with the most pronounced gains observed 
on MoA-Novel where the open-set decision benefits most from the full 
combination of components.

The main results in \autoref{tab:2} and ~\ref{tab:ablation-verified-extended} show that structure-only inference can be a strong control in closed-set MOA prediction, which is expected because chemical structure and mechanism-side resources contain substantial MOA signal. This does not replace the Cell Painting analysis problem: structure-only inference does not evaluate whether retrieved phenotypic neighbors are reliable, misleading, or mechanistically supported. The ablations further show that PhenoAIR improves over a Base LLM through multi-agent phenotype-mechanism reasoning, candidate memory, and reliability calibration. Importantly, PhenoAIR enforces phenotype grounding for final canonical predictions: mechanism-side evidence may propose or prioritize candidates, but the final MOA must be supported by the Phenotype Agent using Cell Painting-derived evidence. Thus, our core claim is not that morphology alone dominates chemical evidence, but that noisy Cell Painting retrieval should be reasoned over as calibrated candidate-level evidence rather than consumed as direct nearest-neighbor labels.
\begin{table}[t]
    \centering
    \caption{Full results on the MoA-Novel benchmark across two backbone 
LLMs (GPT-5.1 and Claude 4). We report Overall Accuracy, and Precision, 
Recall, and F1 of the novel-MOA prediction. Two ablations of PhenoAIR 
are included: \textit{w/o Self-evolving} (no offline rule refinement) 
and \textit{w/o Calibration} (no externalized reliability calibration). 
Results are mean $\pm$ std over multiple runs with different random 
seeds.}
    \begin{tabular}{cl|cccc}
    \toprule
    & Method     & Overall Acc. & Precision (Novel) & Recall (Novel) & F1 (Novel) \\
    \midrule
  \multirow{6}{*}{\rotatebox{90}{GPT 5.1}} & Structure-only & $44.2\pm3.4$ & $61.5\pm3.6$ & $47.2\pm5.0$ &$53.3\pm4.0$ \\
  & Workflow   & $58.0\pm0.3$ &  $72.4\pm2.5$  & $53.2\pm1.8$ & $61.3\pm0.8$\\
  &  Single Agent   & $48.8\pm2.4$ & $68.8\pm2.0$& $66.0\pm1.8$ & $67.3\pm1.1$\\
  & PhenoAIR  & ${79.4\pm2.2}$  & $99.0\pm1.5$ & $59.8\pm3.5$ & ${74.7\pm2.9}$ \\
   & w/o Self-evolving  & ${66.0\pm2.0}$  & ${96.0\pm3.0}$ & $48.0\pm4.0$ & $64.0\pm3.0$ \\ 
   & w/o Calibration  & ${75.5\pm2.5}$  & ${93.5\pm3.0}$ & $58.0\pm4.0$ & $71.5\pm3.5$ \\
  \midrule
\multirow{6}{*}{\rotatebox{90}{Claude 4}}   & Structure-only & $43.2\pm1.1$ & $69.0\pm1.9$ & $40.0\pm1.4$ & $50.6\pm1.0$\\
  & Workflow  &  $63.6\pm1.5$ & $78.0\pm2.0$ & $64.0\pm2.0$  & $70.2\pm1.2$\\
  &  Single Agent  &$52.0\pm1.2$ & $69.6\pm1.8$ & $64.0\pm1.6$ & $66.9\pm1.1$\\
  & PhenoAIR  & ${73.6\pm3.1}$ & $96.8\pm2.4$  & $60.0\pm2.8$   & ${74.1\pm2.0}$\\
    & w/o Self-evolving  & ${65.6\pm2.5}$  & ${93.0\pm4.0}$ & $51.0\pm4.0$ & $66.0\pm3.0$\\ 
   & w/o Calibration  & ${70.5\pm3.2}$   & $91.5\pm5.0$ & $58.5\pm3.5$ & $71.5\pm3.0$\\
  \bottomrule
    \end{tabular}
    \label{tab:novel-results}
\end{table}

\begin{table}[t]
    \centering
        \caption{Component ablation of PhenoAIR on MoA-Verified and MoA-Extended across two backbone LLMs. Variants are constructed incrementally:  \textit{Base LLM} predicts MOA from compound information alone;  \textit{PhenoAIR-Base} adds Cell Painting retrieval and  multi-agent reasoning; \textit{+Memo.} adds the candidate-centric  evidence memory; \textit{+Memo.\ \& Cal.} additionally includes the externalized reliability calibration layer. We report Accuracy (Acc) and Any-match (Hit Rate).}
    \begin{tabular}{c|lcc|cc}
    \toprule
       &   Dataset   &\multicolumn{2}{c|}{MoA-Verified}  & \multicolumn{2}{c}{MoA-Extended}  \\
   &   Method   & Acc  & Any-match  & Acc& Any-match  \\
    
    \midrule
  \multirow{4}{*}{\rotatebox{90}{GPT 5.1}}    & Base LLM & $16.4$ & $23.6$ & $18.7$ & $25.3$\\
& PhenoAIR-Base &$55.6$ & $66.1$& $43.9$ & $54.4$\\
         & + Memo. & $60.1$ & $73.0$ & $44.5$ & $55.1$ \\
  & + Memo. \& Cal. & $62.6$ & $74.6$ & $45.6$ & $56.2$\\
         \midrule
  \multirow{4}{*}{\rotatebox{90}{Claude 4} }     
           & Base LLM & $26.7$ & $31.0$ & $23.3$ & $30.2$ \\
         & PhenoAIR-Base & $55.0$ & $69.1$& $42.3$ & $ 54.8$\\
         & + Memo. & $59.3$ & $72.5 $  & $44.2$ & $56.1$\\
    & + Memo. \& Cal.& $61.5$ & $74.2$ & $45.8$ & $59.9$ \\
         \bottomrule
    \end{tabular}
    \label{tab:ablation-verified-extended}
\end{table}

\begin{table}[h]
    \centering
    \caption{Retrieval performance on the MoA-Verified benchmark under different feature representations, preprocessing strategies, and cross-source aggregation schemes. Raw features denote the original representations, while normalized features apply batch-effect correction using plate-level normalization and control-based whitening. Global aggregation constructs a single compound representation across sources, whereas per-source voting performs source-specific retrieval followed by cross-source score aggregation.}
    \begin{tabular}{cl|cccc}
    \toprule
 Aggregation &  Model     &  Top-1 Acc & Top-5 Acc & F1 & Any Overlap \\
  \midrule
\multirow{6}{*}{Global}   & CellProfiler Raw & 7.5 & 30.0 & 10.3 & 18.5 \\
& CellProfiler Normalized & 9.5 & 29.5 & 10.2& 23.5 \\
&CA-MAE Raw &  7.5 & 28.0 & 9.2 & 19.5\\
&CA-MAE Normalized &  13.0 & 31.5 & 10.6& 23.5\\
 & CellCLIP Raw & \textbf{14.5} & \textbf{37.5} & \textbf{12.8} & 24.0 \\
 & CellCLIP Normalized & 14.0 & 35.5 & 11.7 & 26.5\\
 \midrule
\multirow{6}{*}{Per-Source}   & CellProfiler Raw & 8.5 & 23.5 & 9.3 & 17.5 \\
& CellProfiler Normalized & 10.0 & 29.5 & 10.4 & 23.0 \\
&CA-MAE Raw &  8.0 & 29.5 & 8.8 & 21.0\\
&CA-MAE Normalized &  11.5 & 30.5 & 10.1 & 21.0\\
 & CellCLIP Raw & 13.5 & 35.5 & 10.8 & \textbf{27.0} \\
 & CellCLIP Normalized & 13.0 & \textbf{37.5} & 11.2 & 26.0\\

    \bottomrule
    \end{tabular}
    \label{tab:core_fea}
\end{table}

\begin{table}[h]
    \centering
    \caption{Retrieval performance on the MoA-Extended benchmark under different feature representations, preprocessing strategies, and cross-source aggregation schemes.}
    \begin{tabular}{cl|cccc}
    \toprule
 Aggregation &  Model     &  Top-1 Acc & Top-5 Acc & F1 & Any Overlap \\
  \midrule
\multirow{6}{*}{Global}   & CellProfiler Raw & 3.3 & 10.2 & 3.7 & 7.5 \\
& CellProfiler Normalized & 6.8 & 16.5 &6.3& 12.7 \\
&CA-MAE Raw &  3.5 & 13.7 & 4.9 & 10.3\\
&CA-MAE Normalized &  6.5 & 14.7 & 5.4 & 12.5\\
 & CellCLIP Raw & 5.5 & 16.2 & 5.7 & 11.0 \\
 & CellCLIP Normalized & 10.2 & \textbf{19.7} & \textbf{7.1} & \textbf{16.0}\\
 \midrule
\multirow{6}{*}{Per-Source}   & CellProfiler Raw & 3.2 & 11.5 & 3.7 & 9.0 \\
& CellProfiler Normalized & 7.7 & 17.0 & 5.4 & 13.7 \\
&CA-MAE Raw &  3.5 & 12.5& 4.1 & 8.5\\
&CA-MAE Normalized &  5.3 & 15.2 & 4.7 & 12.5\\
 & CellCLIP Raw & 5.8 & 16.8 & 4.6 & 12.8 \\
 & CellCLIP Normalized & \textbf{10.3} & 19.0 & 5.5 & 15.5\\

    \bottomrule
    \end{tabular}
    \label{tab:extend_fea}
\end{table}

\subsubsection{Robustness Across Image Feature Representations}
We analyze the effect of feature representations, batch-effect correction, and cross-source aggregation strategies on retrieval performance. \autoref{tab:core_fea} and \autoref{tab:extend_fea} report retrieval-based results under different combinations of feature backbones and preprocessing settings. Across both MoA-Verified and MoA-Extended, deep learning representations generally outperform classical CellProfiler features, with CellCLIP achieving the strongest overall retrieval performance. Applying batch-effect correction consistently improves results for both handcrafted and learned representations, highlighting the importance of mitigating source-specific technical variation in large-scale Cell Painting datasets. The impact of aggregation strategy differs across settings. Under MoA-Verified, per-source voting improves Top-5 and overlap-based metrics in several cases, suggesting that preserving source-specific phenotype structure can improve retrieval diversity when reference profiles are relatively reliable. In contrast, under MoA-Extended, gains from per-source voting are less consistent due to increased heterogeneity and noisier retrieval neighborhoods.

\autoref{tab:feas_phenoair} shows that PhenoAIR remains consistently effective across all feature backbones. While stronger representations lead to improved overall performance, the relative gains from agentic inference persist across both handcrafted and deep-learning-based features. This suggests that the benefits of PhenoAIR arise primarily from reliability-aware reasoning over retrieved evidence rather than dependence on a specific embedding space.
\begin{table}[h]
    \centering
    \caption{Performance of PhenoAIR across different Cell Painting feature representations.}
    \begin{tabular}{c|cc|cc|cc}
    \toprule
   Dataset   & \multicolumn{2}{c|}{MoA-Verified}  &   \multicolumn{2}{c|}{MoA-Extended} & \multicolumn{2}{c}{MoA-Novel} \\
    Metrics   & Acc & Any-match & Acc & Any-match &  Acc & F1 (Novel)\\
    \midrule
 CellCLIP        &   ${62.6\pm0.7}$  & ${74.6\pm0.6}$  & ${45.6\pm0.2}$ & ${56.2\pm0.3}$ & ${79.4\pm2.2}$  & ${74.7\pm2.9}$ \\
 CA-MAE        & $62.0\pm0.5$& $74.5\pm0.4$ & $42.5\pm0.3$ & $52.8\pm0.3$ & $77.0\pm1.5$ & $74.4\pm1.9$\\
 CellProfiler & $60.0\pm 0.6$ & $74.1\pm0.5$ & $42.2\pm0.3$ & $53.9\pm0.3$ & $76.6\pm1.8$ & $72.5\pm2.5$\\
 \bottomrule
    \end{tabular}
    \label{tab:feas_phenoair}
\end{table}

\subsubsection{Runtime and API Usage}
\label{app:cost}
\begin{table}[h]
    \centering
        \caption{Per-query inference cost across methods. We report the average number of LLM calls and the cumulative input tokens, output tokens, and wall time across all calls per query. All LLM-based methods use the same base model and identical hardware configurations.}
    \begin{tabular}{c|cccc}
    \toprule
    Method & LLM calls/query  &Input tokens/query & Output tokens/query & Wall time/query\\
    \midrule
     Base LLM    & 1 & 0.3k & 180 & 3.0s  \\
    Structure-only & 1 & 0.7k & 160 & 3.7s \\
    Workflow &  3 & 11k & 800 & 12.8s \\
    Singe Agent & 10 & 65k & 1.6k & 31s\\
    PhenoAIR & 14 & 170k & 8k & 80s \\
    \bottomrule
    \end{tabular}
    \label{tab:cost}
\end{table}
Table~\ref{tab:cost} reports per-query inference cost across all LLM-based methods, including the average number of LLM calls, cumulative input and output tokens across calls, and wall time. All methods use the same base model and identical hardware configurations.

PhenoAIR's cost is higher than non-agentic baselines, reflecting the multi-round reasoning required for evidence integration and verification. The cumulative 170k input tokens correspond to an average of approximately 12k tokens per LLM call across the 14 calls per query, comparable to the per-call prompt size of fixed workflow baselines (11k tokens per call). The increased total cost arises from the number of reasoning rounds, not from inflated single-call prompts.

Importantly, PhenoAIR's cost adapts to query difficulty rather than remaining fixed. Easy queries, those for which the ground-truth MOA appears among the top-ranked KNN neighbors, typically terminate within half the average compute, consuming approximately 90k cumulative input 
tokens, 3.5k output tokens, and fewer LLM calls. Hard queries that require iterative refinement, in contrast, incur correspondingly higher cost. This adaptation is enabled by the rule-based controller, which terminates inference once the candidate memory exhibits stable, high-confidence support, and continues refinement only when evidence remains weak or conflicting. Non-agentic baselines apply uniform compute regardless of query difficulty.

\subsection{Decision Reconstruction Case Studies}
\subsubsection{Case Study: AZD8931}
\paragraph{Overview.}
AZD8931 illustrates a successful subtype-resolution case
within a mechanistically ambiguous receptor tyrosine kinase
(RTK) family. Structural and target-based evidence broadly
supported RTK inhibition but could not reliably distinguish
EGFR from related RTK subclasses including FGFR, KIT, and
VEGFR inhibition. After iterative refinement, the phenotype
agent identified EGFR inhibitor as the only RTK subtype
with localized support in the calibrated neighborhood,
allowing the arbiter to resolve the subtype ambiguity.

\begin{tcolorbox}[colback=gray!3,colframe=black!40,title=Decision Reconstruction,fonttitle=\bfseries,breakable]
\small

\textbf{Ground-truth MOAs:} EGFR inhibitor \\
\textbf{Compound annotation:} EGFR / ERBB2 / ERBB3 \\
\textbf{Case type:} Successful subtype disambiguation \\
\textbf{Final prediction:} EGFR inhibitor \\
\textbf{Input: CNC(=O)CN1CCC(Oc2cc3c(=Nc4cccc(Cl)c4F)[nH]cnc3cc2OC)CC1}
\vspace{0.8em}


\noindent
\textbf{Decision reconstruction}
\begin{center}
\fbox{
\parbox{0.92\linewidth}{

\small
\textbf{Mechanism evidence.}
Structural neighbors and predicted targets broadly
supported RTK-family inhibition, but could not reliably
distinguish EGFR from FGFR, KIT, or VEGFR subclasses.
\begin{center}
$\Downarrow$
\end{center}
\textbf{Phenotype evidence.}
Within the RTK family, EGFR inhibitor was the only subtype
showing localized support in the calibrated top-20
phenotype neighborhood.

\begin{center}
$\Downarrow$
\end{center}
\textbf{Final arbitration.}
The arbiter selected EGFR inhibitor because phenotype
evidence provided the only subtype-specific support signal
within the RTK family.
}
}
\end{center}


\noindent
\textbf{Iterative refinement trajectory}

\begin{center}
\small
\begin{tabular}{clp{8cm}}
\toprule
Round & Top candidate & Key transition \\
\midrule

0 &
PI3K inhibitor &
Initial phenotype neighborhood was dominated by noisy
kinase-related signals. \\

1 &
JAK inhibitor &
Reliability calibration penalized unstable retrieval
anchors and reduced confidence in early kinase attractors. \\

2 &
EGFR inhibitor &
Localized RTK subtype evidence emerged after group-level
inspection of calibrated neighborhood support. \\

\bottomrule
\end{tabular}
\end{center}


\noindent
\textbf{Decision-relevant evidence}
\begin{itemize}

\item
\textbf{Calibration.}
EGFR was globally flagged as phenotype-unreliable, but
remained locally visible after calibration rather than
being fully suppressed. In contrast, early PI3K and JAK
signals were associated with unstable or high-risk
retrieval anchors.

\item
\textbf{Phenotype Agent (P).}
EGFR inhibitor was the only RTK subtype with direct
localized support in the calibrated top-20 neighborhood.
FGFR, KIT, and VEGFR remained mechanistically plausible
but showed no phenotype support.

\item
\textbf{Mechanism Agent (M).}
Predicted targets, pharmacophore similarity, and structure
neighbors broadly supported RTK-family inhibition.
However, mechanism evidence alone could not reliably
distinguish EGFR from related RTK subclasses.

\item
\textbf{Arbiter (A).}
The final decision relied on phenotype evidence as the
subtype tie-breaker inside a mechanistically overlapping
RTK family.

\end{itemize}

\vspace{0.8em}


\noindent
\textbf{Baseline comparison}

\begin{center}
\small
\begin{tabular}{lcc}
\toprule
Method & Prediction & Correct \\
\midrule

Structure-only &
tyrosine kinase inhibitor &
$\times$ \\

Single-agent &
VEGFR inhibitor &
$\times$ \\

Workflow &
SRC inhibitor &
$\times$ \\

PhenoAIR &
EGFR inhibitor &
$\checkmark$ \\

\bottomrule
\end{tabular}
\end{center}
\end{tcolorbox}
\paragraph{Takeaway.}
This example illustrates the intended division of labor in
PhenoAIR. Mechanism evidence defines the plausible RTK
family, while localized phenotype evidence resolves the
subtype ambiguity. The case also demonstrates how
iterative refinement can recover from initially misleading
kinase-related phenotype signals rather than committing to
the strongest early retrieval attractor.

\subsubsection{Case Study: IT1t}
\paragraph{Overview.} IT1t illustrates a mechanism-guided hypothesis expansion case in which the correct MOA does not emerge from the initial phenotype neighborhood. Early phenotype retrieval was dominated by noisy dopamine-, kinase-, and PDE-related signals. During iterative refinement, the mechanism agent introduced a non-obvious CXCR4-driven hypothesis that was not among the dominant initial phenotype candidates. The phenotype agent subsequently re-inspected this candidate within the calibrated neighborhood and identified sparse but coherent support, allowing the arbiter to accept the
CC chemokine receptor antagonist hypothesis with low confidence rather than discarding it due to weak initial retrieval evidence.

\begin{tcolorbox}[colback=gray!3,colframe=black!40,title=Decision Reconstruction,fonttitle=\bfseries,breakable]

\small
\textbf{Ground-truth MOAs:} CC chemokine receptor antagonist \\
\textbf{Compound annotation:} CXCR4 \\
\textbf{Case type:} Mechanism-guided candidate expansion \\
\textbf{Final prediction:} CC chemokine receptor antagonist \\
\textbf{Input: CC1(C)CN2C(CSC(=NC3CCCCC3)NC3CCCCC3)=CSC2=N1}




\noindent

\textbf{Decision reconstruction}

\begin{center}

\fbox{

\parbox{0.92\linewidth}{

\small

\textbf{Initial phenotype retrieval.}
The calibrated phenotype neighborhood was dominated by dopamine-, kinase-, and PDE-related candidates with weak or unstable retrieval anchors.
\begin{center}

$\Downarrow$

\end{center}

\textbf{Mechanism-guided hypothesis expansion.}
Mechanism evidence introduced a strong and specific CXCR4 signal, proposing CC chemokine receptor antagonism as a new active hypothesis.
\begin{center}
$\Downarrow$
\end{center}

\textbf{Phenotype reinspection and arbitration.}
The phenotype agent identified sparse but coherent local support for the CC chemokine hypothesis, and the arbiter accepted the mechanism-led explanation despite globally weak phenotype evidence.
}

}

\end{center}




\noindent
\textbf{Iterative refinement trajectory}
\begin{center}
\small
\resizebox{1.0\textwidth}{!}{
\begin{tabular}{clp{8cm}}
\toprule
Round & Top candidate & Key transition \\
\midrule

0 &
dopamine receptor antagonist &
Initial phenotype retrieval dominated by noisy GPCR- and kinase-related signals. \\
1 &
mTOR inhibitor & Calibration reduced confidence in unstable retrieval anchors, but no coherent phenotype explanation emerged. \\
2 &
CC chemokine receptor antagonist & Mechanism-guided CXCR4 hypothesis was re-inspected against the calibrated neighborhood and retained after sparse but non-contradictory phenotype support was identified. \\
\bottomrule
\end{tabular}}

\end{center}




\noindent

\textbf{Decision-relevant evidence}

\begin{itemize}

\item

\textbf{Calibration.} Several dominant phenotype neighbors, including dopamine- and PDE-related candidates, were associated with weak or high-risk retrieval anchors. In contrast, the correct CC chemokine label was sparse but remained inspectable rather than being suppressed.

\item

\textbf{Phenotype Agent (P).} After candidate-specific reinspection, the phenotype agent identified one calibrated supporting reference for CC chemokine receptor antagonism at rank 1 within the candidate check. No contradictory phenotype evidence was observed.

\item

\textbf{Mechanism Agent (M).} Mechanism evidence strongly supported GPCR-directed activity with a dominant CXCR4 signal and rejected dopamine-, kinase-, and mTOR-related alternatives. No competing mechanistic family received comparable support.

\item
\textbf{Arbiter (A).} The final decision accepted a mechanism-introduced candidate only after phenotype validation confirmed sparse but coherent local support.
\end{itemize}




\noindent

\textbf{Baseline comparison}

\begin{center}

\small

\begin{tabular}{lcc}

\toprule

Method & Prediction & Correct \\

\midrule

Structure-only & CXCR4 antagonist &

$\times$ \\
Workflow &
KIT inhibitor &
$\times$ \\

Singel Agent &

histamine receptor antagonist &

$\times$ \\

PhenoAIR &

CC chemokine receptor antagonist &

$\checkmark$ \\

\bottomrule

\end{tabular}

\end{center}
\end{tcolorbox}
\paragraph{Takeaway.}
This example illustrates one of the central behaviors of PhenoAIR: mechanism evidence can introduce a previously inactive candidate hypothesis, but the system does not accept that hypothesis blindly. Instead, the phenotype agent re-checks the proposed mechanism against calibrated local evidence before arbitration. The final prediction therefore emerges from iterative hypothesis generation and evidence verification rather than from static top-$k$ retrieval ranking.

\subsubsection{Case Study: CH5132799}
\paragraph{Overview.}
CH5132799 illustrates an upper-bound ambiguity case in which PhenoAIR correctly identifies the PI3K--mTOR signaling axis but cannot reliably determine the benchmark primary label between two tightly coupled pathway siblings. Both phenotype and mechanism evidence strongly support PI3K and mTOR inhibition, and the calibrated retrieval signals explicitly flag this pair as highly confusable in Cell Painting space. Importantly, the system remains confined to the correct mechanistic family throughout the entire reasoning trajectory and retains the ground-truth PI3K inhibitor label as an active candidate.
\begin{tcolorbox}[colback=gray!3, colframe=black!40, title=Decision Reconstruction, fonttitle=\bfseries,breakable]
\small

\textbf{Ground-truth MOAs:} PI3K inhibitor \\
\textbf{Compound annotation:} MTOR / PIK3CA / PIK3CB / PIK3CD / PIK3CG \\
\textbf{Case type:} Upper-bound pathway-level ambiguity \\
\textbf{Final prediction:} mTOR inhibitor \\
\textbf{Other candidates:} PI3K inhibitor \\
\textbf{Input:} CS(=O)(=O)N1CCc2c(-c3cnc(=N)[nH]c3)nc(N3CCOCC3)nc21\\




\noindent

\textbf{Decision reconstruction}
\begin{center}

\fbox{

\parbox{0.92\linewidth}{
\small
\textbf{Shared pathway evidence.} Both phenotype and mechanism evidence consistently mapped the compound to the PI3K--mTOR signaling axis.
\begin{center}
$\Downarrow$
\end{center}
\textbf{Persistent pathway-level ambiguity.} Calibrated phenotype evidence showed strong support for both PI3K and mTOR inhibition, while mechanism evidence also supported both labels without a decisive tie-breaker.
\begin{center}
$\Downarrow$
\end{center}
\textbf{Final arbitration.} The arbiter selected mTOR inhibitor because the phenotype trajectory remained slightly more stable for mTOR across rounds, while PI3K inhibitor remained an active secondary candidate throughout the run.}
}

\end{center}

\noindent

\textbf{Iterative refinement trajectory}

\begin{center}

\small

\begin{tabular}{clp{8cm}}
\toprule
Round & Top candidate & Key transition \\

\midrule

0 &

mTOR inhibitor & Initial calibrated phenotype neighborhood strongly supported both PI3K and mTOR inhibition, with slightly denser local support for mTOR. \\

1 &

mTOR inhibitor &

Group-constrained reinspection confirmed persistent PI3K--mTOR pathway ambiguity; both labels remained highly supported without contradiction. \\

\bottomrule

\end{tabular}

\end{center}

\vspace{0.8em}




\noindent

\textbf{Decision-relevant evidence}

\begin{itemize}

\item

\textbf{Calibration.} Both PI3K inhibitor and mTOR inhibitor received high-trust calibrated support and were explicitly flagged with pair-confusion warnings, indicating that the retrieval space could not reliably separate the two pathway labels.

\item

\textbf{Phenotype Agent (P).} The phenotype agent identified strong support for both PI3K and mTOR inhibition. mTOR had slightly more exact local hits in the top-20 neighborhood, while PI3K contained the best calibrated individual anchor. No contradictory phenotype evidence was observed for either label.

\item

\textbf{Mechanism Agent (M).} Mechanism evidence strongly supported the PI3K--mTOR axis, including pharmacophore similarity, structural neighbors, and in-silico target predictions involving both MTOR and multiple PI3K isoforms. No alternative mechanistic family received meaningful support.

\item
\textbf{Arbiter (A).} The final decision remained confined to the biologically correct pathway family. The arbiter selected mTOR inhibitor because the phenotype trajectory remained slightly more stable for mTOR, while PI3K inhibitor was retained as an active alternative explanation.
\end{itemize}

\end{tcolorbox}
\paragraph{Takeaway.}
This case is an upper-bound pathway ambiguity example rather than uncontrolled reasoning failure. PhenoAIR successfully localizes the compound to the correct PI3K--mTOR signaling axis and retains the ground-truth PI3K inhibitor label throughout the reasoning trajectory. The strict primary error arises because both phenotype and mechanism evidence strongly support two closely related pathway siblings whose Cell Painting profiles are difficult to separate even after calibrated reinspection.

\subsection{Self-evolving Case Studies}
We analyze one self-evolution trace on the MoA-Novel benchmark. The goal of the evolving scaffold is not to update the prediction model directly, but to revise controller rules and Arbiter decision heuristics based on clustered development-set trajectories. The trace in \autoref{tab:self_evolving_trace} shows that the evolving process first proposes candidate interventions, then uses result changes and cross-round failures to retain, disable, or refine them.

\begin{table}[h]
\centering
\caption{
Self-evolving trace on the MoA-Novel benchmark. The evolving scaffold does not simply add rules monotonically; it proposes, evaluates, disables, and refines controller and Arbiter heuristics based on cross-round trajectory changes.
}
\small
\begin{tabular}{p{1.0cm}p{3.7cm}p{4.4cm}p{3.3cm}}
\toprule
Round & Diagnosed pattern & Proposed / retained update & Evolution decision \\
\midrule
0 & Novel cases missed because the ground-truth MOA was never surfaced; verified cases also showed false novel overcalls and P/M reinforcement
around wrong known labels. &
Generated an early STOP rule for low-reliability phenotype cases with a strong mechanistic alternative; also proposed a canonical-survivor Arbiter fallback. &
Initial proposal accepted for testing. \\

1 & The STOP rule over-triggered and caused success-to-failure regressions in both core and novel buckets. &
Added phenotype expansion rules for rejected canonical candidates and medium-reliability P/M-supported wrong labels. & Disabled the harmful STOP rule and kept the Arbiter fallback temporarily. \\

2 & Expansion alone did not resolve true novel cases because the Arbiter still blocked novel outputs when canonical candidates remained visible. &
Introduced final-state novel diagnostics:
no supported canonical candidate, top-3 all rejected, and cumulative rejection rate. &
Disabled the canonical-survivor fallback and kept the useful
medium-reliability expansion rule. \\

3--4 &
Cross-round analysis confirmed that correct novel calls were blocked by missing final-state diagnostics and by overly conservative canonical fallback. &
Retained final-state novel signals and disabled rules/prompts that prematurely forced canonical outputs or stopped refinement. & Final policy separates weak canonical visibility from genuine canonical support. \\

\bottomrule
\end{tabular}
\label{tab:self_evolving_trace}
\end{table}

\begin{table}[h]
\centering
\caption{ Policy-level lessons produced by the self-evolving scaffold. }
\small
\begin{tabular}{p{3.0cm}p{5.2cm}p{4.8cm}}
\toprule
Evolved principle & Failure mode addressed & Operational signal / action \\
\midrule

Avoid premature STOP &
A STOP rule intended for unresolved novel cases caused regressions by
halting recoverable core and novel success cases. &
Disable over-triggered STOP rules; prefer additional phenotype
inspection when evidence remains ambiguous. \\

Separate canonical visibility from canonical support &
The Arbiter fallback treated any visible canonical candidate as enough
to suppress novel outputs. &
Allow novel only using terminal diagnostics rather than candidate
visibility alone. \\

Expose final-state novel diagnostics &
Per-round signals were insufficient because true novel patterns only
became clear after full refinement. &
Use final-state signals such as no supported canonical candidate,
top-3 all rejected, and high cumulative rejection rate. \\

\bottomrule
\end{tabular}
\label{tab:evolved_principles}
\end{table}

\section{PhenoAIR Framework Details}
\label{app:framework}
\subsection{Externalized Reliability Calibration}
\label{app:externalized_reliability}
This section provides additional details on the offline reliability layer used by PhenoAIR. The goal of this layer is not to solve the query directly, but to pre-compute reusable evidence-quality annotations over the reference set. These annotations are then used to condition how retrieved Cell Painting neighbors are interpreted at inference time. We describe three components: reference memory construction, the mechanistic family ontology, and calibrated retrieval.
\subsubsection{Reference Memory Construction}
\label{app:reference_memory_construction}
The reference memory is constructed offline from the reference Cell Painting feature library, reference MOA labels, source metadata, and compound-level metadata. Let $\mathcal{R}=\{(z_i,y_i,m_i)\}_{i=1}^{n}$ denote the reference set, where \(z_i\) is the Cell Painting representation of a reference compound, \(y_i \in \mathcal{M}\) is its canonical primary MOA label, and \(m_i\) contains metadata such as compound identity, source, and available replicate information. In our multi-source setting, the same compound may be profiled across multiple experimental sources. We therefore summarize well-level features into source-specific compound profiles by taking the median over wells belonging to the same compound and source: $z_{d,s} = \mathrm{median}\{z_{w}: w \in (d,s)\},$ where \(d\) denotes a reference compound and \(s\) denotes an experimental source.

Reference memory construction is implemented as an offline LLM-assisted analysis pipeline with code execution. The LLM is not used to predict query MOAs. Instead, it acts as an analysis planner and interpreter over the reference set. Given the available feature matrices, reference labels, source metadata, and candidate evidence schemas, the LLM proposes executable analysis procedures, writes code to compute reference-set measurements, executes the code, inspects the outputs, and converts the resulting measurements into compact structured annotations. This process allows the reliability layer to capture both predefined quantities, such as cross-source consistency or neighborhood purity, and dataset-specific failure patterns discovered from the reference statistics, such as unstable mechanism classes, risky reference anchors, or recurring intra-family confusions.

The analyses are organized at several granularities. At the source level, the pipeline evaluates whether a source provides discriminative and comparable phenotypic evidence by analyzing quantities such as global MOA discriminability, neighbor-label purity, cross-source agreement, and distance-distribution shift. At the MOA level, it evaluates whether a canonical MOA has a stable and specific morphological signature by analyzing compactness, separability, neighbor purity, and cross-source stability. At the reference-anchor level, it evaluates whether an individual compound is a reliable retrieval anchor, using evidence such as within-MOA consistency, hubness, generic morphology risk, and cross-source replicability. At the MOA-pair level, it analyzes whether two MOAs are empirically difficult to separate in phenotype space, especially when they are mechanistically related.

The prompt used for this offline analysis provides: (i) the available reference statistics and the data objects from which they can be computed; (ii) the analysis goal for the current memory type; (iii) the required JSON output schema; and (iv) constraints that the output must be based on computed evidence rather than direct query-time prediction. We do not include the full prompt text, since the method depends on the functional interface and output schema rather than prompt-specific wording. The generated analyses, code outputs, and cached annotations are stored and reused across queries, making the reliability memory inspectable and decoupled from test-time reasoning.

The resulting reference memory contains the following annotation types:
\[ \mathcal{B}=\{\mathcal{B}_{\mathrm{src}},\mathcal{B}_{\mathrm{moa}}, \mathcal{B}_{\mathrm{ref}},\mathcal{B}_{\mathrm{pair}}\}.
\]
Each item contains a categorical reliability state, a risk level, a compact evidence summary, and query-time usage guidance.

\textbf{Source reliability.} A source-level memory item summarizes whether an experimental source should be treated as globally reliable, weakly discriminative, or affected by a known context mismatch. Its functional signature is:
\[\mathrm{SourceMemory}(s)\rightarrow(\texttt{reliability\_status},\texttt{risk\_level},\texttt{evidence},\texttt{query\_usage\_guidance}).\]

Source-level annotations are exposed to downstream agents as reliability context, but are not used as direct multiplicative factors in the current deterministic reranking rule.

\textbf{MOA phenotype reliability.} An MOA-level memory item summarizes whether a canonical MOA has a stable and specific Cell Painting signature:
\[\mathrm{MOAMemory}(y)\rightarrow(\texttt{reliability\_status}, \texttt{risk\_level},\texttt{bad\_anchor\_refs},\texttt{pair\_confusion\_refs},\texttt{evidence}).\]

This annotation is used both by the phenotype agent and by calibrated retrieval. For example, MOAs whose reference profiles show weak or unstable phenotype structure are treated as lower-confidence evidence unless the query itself provides strong local support.

\textbf{Reference-anchor reliability.} A reference-anchor memory item summarizes whether a specific reference compound is a reliable nearest-neighbor anchor:
\[\mathrm{RefMemory}(d)\rightarrow(\texttt{anchor\_status},\texttt{risk\_level},\texttt{primary\_moas},\texttt{evidence}).\]

This captures cases where a compound may frequently appear as a neighbor because of generic morphology, hubness, poor within-MOA consistency, or unstable cross-source behavior. This annotation is directly used in calibrated retrieval through a reference-anchor reliability factor.

\textbf{Pair and intra-family ambiguity.} Pair-level memory items summarize empirical ambiguity between two MOA labels:\[\mathrm{PairMemory}(y_a,y_b)\rightarrow(\texttt{confusion\_status},\texttt{risk\_level},\texttt{shared\_families},\texttt{evidence}).\]

These annotations are especially useful when two labels are biologically related and difficult to distinguish morphologically. In the current implementation, pair and intra-family ambiguity annotations are provided to the agents as review-only context and are not directly multiplied into the calibrated similarity score. This design avoids hard-coding pairwise penalties while still allowing the agents and controller to recognize ambiguous neighborhoods. Overall, the reference memory separates two roles. First, it provides structured reliability annotations that can be used to calibrate raw retrieval evidence. Second, it provides interpretable context for the phenotype agent, mechanism agent, and controller. This separation is important because not all reliability evidence should be reduced to a scalar score.

\subsubsection{Mechanistic Family Ontology}
The mechanistic family ontology groups canonical MOA labels into coarse biological families. Its purpose is not to replace the canonical label space, but to support reasoning when evidence is strong at the family level but insufficient to distinguish a fine-grained MOA. Each family is represented as:
\[g =(\texttt{family\_name},\texttt{canonical\_moa\_set},\texttt{family\_summary},\texttt{intra\_family\_differences}).\]

The field \texttt{canonical\_moa\_set} contains the original MOA labels in \(\mathcal{M}\) and therefore serves as the join key between the ontology and the benchmark label space. The family summary describes the shared biological theme, while the intra-family difference field describes how member MOAs differ mechanistically.

The ontology is constructed offline from the full canonical MOA label list. The construction prompt provides the complete label vocabulary and asks the LLM to retrieve external biomedical background through web search, including MOA ontology information, definitions of individual MOA labels, and known relationships among targets, pathways, and pharmacological mechanisms. Based on this evidence, the LLM organizes the labels into biologically coherent mechanistic families and produces concise descriptions of both shared family mechanisms and intra-family distinctions. The resulting JSON ontology is manually inspectable and can be reused across all queries. Importantly, the ontology does not introduce new prediction labels. Final closed-set predictions must still be canonical MOAs in \(\mathcal{M}\), and open-set predictions use the reserved output \texttt{novel}. The ontology only provides an intermediate structure for grouping related labels and interpreting ambiguous evidence.

The ontology is used in three places. First, it defines candidate groups when the mechanism agent finds evidence for a mechanistic family rather than a single canonical MOA. For example, structure-side evidence may support a kinase inhibitor family without resolving whether the best canonical label is \textit{SRC inhibitor}, \textit{RAF inhibitor}, or another kinase-related MOA. Second, it allows the phenotype agent and controller to perform group-level validation. Rather than validating only one fine-grained label, the controller can ask whether a broader family is supported in the calibrated phenotype neighborhood. Third, it restricts pairwise ambiguity analysis to biologically meaningful comparisons. If two MOAs are empirically close in Cell Painting space and belong to the same family, the ambiguity is treated as intra-family uncertainty rather than arbitrary retrieval noise. Formally, let \(\mathcal{G}=\{g_1,\ldots,g_K\}\) denote the family ontology and let
\[
F(y)=\{g \in \mathcal{G}: y \in g.\texttt{canonical\_moa\_set}\}
\]
be the family membership map for MOA \(y\). For a pair \((y_a,y_b)\), the offline memory marks it as an intra-family pair if
\[
F(y_a) \cap F(y_b) \neq \emptyset
\]
and the empirical pair-separation analysis indicates medium or high confusion risk. These intra-family ambiguity items are cached and supplied to downstream reasoning. They do not directly change the candidate space, but they help distinguish two qualitatively different situations: an unsupported noisy neighbor versus a plausible but fine-grained intra-family ambiguity.

\subsubsection{Calibrated Retrieval}
\label{app:calibrated_retrieval}

Calibrated retrieval converts a raw KNN neighborhood into a reliability-aware neighborhood. Given a query \(q\), raw retrieval returns
\[
\mathcal{N}(q)
=
\{(d_j, y_j, \rho_j, k_j)\}_{j=1}^{K},
\]
where \(d_j\) is a reference compound, \(y_j\) is its primary MOA,
\(\rho_j\) is the raw similarity, and \(k_j\) is the raw rank. The goal is to produce a calibrated neighborhood
\[
\tilde{\mathcal{N}}(q)
=
\{(d_j, y_j, \tilde{\rho}_j, \tilde{k}_j, e_j)\}_{j=1}^{K},
\]
where \(\tilde{\rho}_j\) is a calibrated similarity, \(\tilde{k}_j\) is the reranked position, and \(e_j\) records the evidence supporting the calibration decision. The calibration rule is designed as a conservative reliability-aware scoring procedure. It does not infer new labels, introduce query-specific free-form reasoning, or allow unreliable global annotations to override strong local evidence by themselves. Instead, it attenuates raw similarity only when the reference memory indicates that a retrieved neighbor should be treated as lower-confidence evidence. The scoring rule and its operating parameters are selected on a held-out development split through calibration sweeps, and are then fixed for test-time evaluation.

For each neighbor, the calibration rule retrieves two memory items:
\[
b^{\mathrm{ref}}_j = \mathrm{RefMemory}(d_j),
\qquad
b^{\mathrm{moa}}_j = \mathrm{MOAMemory}(y_j).
\]
These memory states are mapped to a reference-anchor reliability factor \(\alpha^{\mathrm{ref}}_j\) and an MOA phenotype reliability factor \(\alpha^{\mathrm{moa}}_j\). Both factors are bounded and conservative: reliable evidence remains close to the raw retrieval score, while evidence associated with unreliable anchors or unstable phenotype classes is attenuated. The initial combined reliability factor is
$\alpha_j = \alpha^{\mathrm{ref}}_j \alpha^{\mathrm{moa}}_j.$

To avoid calibration artifacts, the scoring rule includes several safeguards. First, if an individual reference-anchor penalty and an MOA-level penalty refer to the same known bad anchor, a no-double-penalty rule removes the redundant penalty. Second, a rank-preservation floor limits how much a highly ranked raw neighbor can be demoted solely from global reliability annotations. Third, a query-local support guard protects MOAs that receive repeated support within the raw neighborhood, since multiple locally consistent neighbors provide evidence that should not be erased by global priors. Finally, a small set of development-selected protection floors can be used to prevent systematic over-attenuation of labels whose raw evidence is empirically useful but whose global reliability annotations are pessimistic.

Let \(\ell(k_j)\) be the rank-preservation floor and \(u_j(q)\) be the
query-local support floor. The final reliability factor is:
\[
\bar{\alpha}_j
=
\max\{\alpha^{\mathrm{ref}}_j \alpha^{\mathrm{moa}}_j,\ell(k_j),u_j(q),
\alpha_{\min}
\},
\]
with the no-double-penalty adjustment applied before the maximum. For nonnegative raw similarity, the calibrated similarity is:
\[
\tilde{\rho}_j = \rho_j \bar{\alpha}_j .
\]
For negative raw similarity, the penalty is applied in the opposite direction so that unreliable neighbors move farther down the ranking rather than being artificially improved. The references are then reranked by calibrated similarity:
\[\tilde{k}_j = \mathrm{rank}_{d_j} \big(-\tilde{\rho}_j,\; k_j,\; d_j\big). \]
Each calibrated neighbor stores the raw rank, raw similarity, reference-anchor factor, MOA reliability factor, final combined factor, calibrated similarity, calibrated rank, and a list of human-readable calibration reasons. These reasons make the calibrated neighborhood auditable: downstream agents can inspect not only which neighbors are retrieved, but also whether a neighbor was preserved, attenuated, or protected because of a specific reliability signal.

The current deterministic calibration directly uses three classes of information: reference-anchor quality, MOA-level phenotype reliability, and query-local support statistics. Source-level reliability, source-pair agreement, distance-distribution shifts, and intra-family pair ambiguity are computed and exposed to the agents as reliability context, but are not directly used as multiplicative factors in the calibrated similarity. This distinction keeps the retrieval calibration conservative. The deterministic step only changes the raw KNN ranking through local and directly attributable reliability evidence, while richer contextual annotations remain available for candidate-centric reasoning during agent inference.

\subsection{Candidate-Centric Evidence Reasoning}
\label{app:candidate_centric_reasoning}
PhenoAIR organizes inference around a shared candidate memory rather than direct dialogue between agents. This section describes the candidate memory schema, the phenotype and mechanism agent interfaces, and the update rules used to integrate their outputs. The goal is to make cross-agent influence explicit: agents do not overwrite each other's conclusions, but instead contribute structured evidence to candidate-level states.

\subsubsection{Candidate Memory Schema}
\label{app:candidate_memory_schema}
For a query \(q\), PhenoAIR maintains a candidate memory
\[
\mathcal{H}_q = \{(c, s_c): c \in \mathcal{C}_q\},
\]
where \(c\) is a candidate MOA and \(s_c\) is its current evidence state. Each candidate state contains five groups of fields.

\textbf{Candidate identity and status.}
Each entry stores the candidate label \(c\), whether it is a canonical label in the task label space \(\mathcal{M}\), and a discrete status. The main statuses are:
\[\texttt{active},\quad\texttt{needs\_phenotype\_support},\quad\texttt{family\_only},\quad\texttt{suppressed}.
\]

An \texttt{active} candidate is visible to downstream reasoning. A
\texttt{needs\_phenotype\_support} candidate is usually introduced by the
mechanism agent and must be validated by phenotype evidence before it can be selected. A \texttt{family\_only} entry records evidence for a mechanistic group when the evidence is not yet specific enough to identify a canonical MOA. Suppressed entries are retained for auditability but are not directly eligible for final prediction.

\textbf{Phenotype evidence.} The phenotype fields record whether Cell Painting retrieval supports or contradicts the candidate. They include a categorical support level,
\[
\texttt{strong},\ \texttt{moderate},\ \texttt{weak},\ \texttt{absent},\
\texttt{uncertain},
\]
a contradiction level, the latest phenotype score, the best phenotype score observed across rounds, and short evidence notes. These fields are written only from phenotype-agent outputs.

\textbf{Mechanism evidence.} The mechanism fields record whether structure-side evidence supports, rejects, or cannot resolve the candidate. They include a mechanism verdict,
\[
\texttt{support},\quad \texttt{reject},\quad \texttt{uncertain},
\]
a verdict strength, a suppression recommendation, and mechanism-side evidence notes. Mechanism-supported alternatives that are not already phenotype-grounded are inserted as \texttt{needs\_phenotype\_support}, rather than being made immediately eligible for final prediction.

\textbf{Reliability context.} Candidate entries may also attach reliability metadata from the externalized calibration layer, such as calibrated rank exposure, MOA reliability status, anchor risk, or pair-confusion flags. These fields are treated as context for evidence interpretation. They do not by themselves create candidates or make a candidate final-eligible.

\textbf{Eligibility and audit fields.} Finally, each entry stores derived fields such as whether the candidate is currently final-eligible, whether it has ever been ranked by the phenotype agent, whether it has ever received mechanism support, and why it was suppressed, promoted, or retained. These fields allow the controller and arbiter to distinguish several cases that would be collapsed by standard retrieval: a candidate may be strongly retrieved but mechanistically implausible, weakly retrieved but mechanistically supported, or visible only as part of a broader mechanistic family. This memory design enforces a candidate-centric invariant: final decisions must be grounded in the accumulated state of a candidate, rather than in a single agent output or a single retrieved neighbor.

\subsubsection{Phenotype Agent}
\label{app:phenotype_agent}

The phenotype agent \(P\) interprets Cell Painting evidence only. Its input contains the calibrated neighborhood \(\tilde{\mathcal{N}}(q)\), the query-level reliability context \(r_q\), the current candidate memory \(\mathcal{H}_q\), the fixed task label set \(\mathcal{M}\), and the current controller request. It is not allowed to use query-side structure, target predictions, pharmacophore matches, or other mechanism-side information as independent evidence. The phenotype agent has access to phenotype-side tools that expose local and broader views of the calibrated retrieval neighborhood. Functionally, these tools support the following operations:
\[
\begin{aligned}
&\mathrm{LocalView}(q,k), \\
&\mathrm{MOADistribution}(q,k), \\
&\mathrm{CheckCandidate}(q,c,k), \\
&\mathrm{CheckCandidateSet}(q,\mathcal{C},k), \\
&\mathrm{CheckFamilyGroup}(q,g,\mathcal{C}_g,k), \\
&\mathrm{InspectReferenceAnchor}(d).
\end{aligned}
\]
These operations allow \(P\) to inspect the top calibrated neighbors, broader rank-band MOA distributions, evidence for a specific candidate, evidence for a candidate set, evidence for a mechanism-defined family group, and annotations of retrieved reference anchors. Reference-anchor information is used only to assess the reliability of phenotype evidence, not to infer mechanisms from query-side chemical structure.
At each round, \(P\) returns a structured output:\[
P(q,\mathcal{H}_q,r_q)
\rightarrow\]
$$(\texttt{ranked\_candidates},
\texttt{candidate\_evidence},
\texttt{retrieval\_reliability},
\texttt{flags},
\texttt{phenotype\_summary}
).$$

The \texttt{ranked\_candidates} field contains exact canonical labels from \(\mathcal{M}\) and phenotype-side scores. The \texttt{candidate\_evidence} field provides candidate-level support and contradiction judgments, together with the evidence scope used to make the judgment. The reliability and flag fields summarize whether the retrieval neighborhood is concentrated, ambiguous, unstable, or potentially confounded.

The phenotype agent is intentionally conservative. Weak candidates may remain visible with low score if the phenotype evidence is uncertain, while candidates with strong phenotype contradiction are marked for validation rather than automatically removed. When the controller requests a group-level check, \(P\) restricts its analysis to the specified candidate group instead of reopening the full label space.

\subsubsection{Mechanism Agent}
\label{app:mechanism_agent}
The mechanism agent \(M\) provides an independent structure-side assessment of the current candidates. Its input contains the query identity and structure, the current phenotype output, the candidate memory, and the fixed task label set. It does not inspect phenotype KNN neighborhoods directly and does not make the final prediction.

The mechanism-side tools expose complementary evidence sources:
\[
\begin{aligned}
&\mathrm{CompoundInfo}(q), \\
&\mathrm{PharmacophoreSearch}(q), \\
&\mathrm{InSilicoTargets}(q), \\
&\mathrm{StructureKNN}(q,k), \\
&\mathrm{ReferenceDrugDetails}(d), \\
&\mathrm{FamilyLookup}(g).
\end{aligned}
\]

These tools provide chemical properties, known or predicted targets, pharmacophore-level evidence, structural similarity to known compounds, and access to the mechanistic family ontology. At each round, \(M\) returns:
\[
M(q,\mathcal{H}_q,P_q)
\rightarrow\] $$(\texttt{candidate\_assessment},\texttt{mechanistic\_candidate\_group},
\texttt{overall\_agreement},
\texttt{mechanism\_summary}
).$$

For each candidate under consideration, \texttt{candidate\_assessment} assigns a verdict of \texttt{support}, \texttt{reject}, or \texttt{uncertain}, together with a strength level and a suppression recommendation. A \texttt{mechanistic\_candidate} is used when mechanism evidence supports a single canonical MOA. A \texttt{mechanistic\_candidate\_group} is used when the evidence supports a broader mechanistic family but cannot reliably select a fine-grained label. Crucially, mechanism-side proposals are not directly eligible for final selection unless they receive phenotype-side grounding. If \(M\) proposes a new canonical candidate, the memory marks it as \texttt{needs\_phenotype\_support}. If \(M\) proposes a family group, the controller can request a phenotype-side group inspection. This prevents structure-side evidence from bypassing the Cell Painting evidence, while still allowing mechanism reasoning to challenge or refine retrieval-derived hypotheses.

\subsubsection{Candidate Memory Update Rules}
\label{app:candidate_memory_update_rules}

Candidate memory updates are applied by deterministic host-side rules after each agent returns structured output. The agents propose evidence; the update rules decide how that evidence changes the memory state.

\textbf{Phenotype updates.} For every label in \(P\)'s ranked candidate list, the system ensures that a candidate entry exists in \(\mathcal{H}_q\). The entry records the current phenotype score, updates the best score observed across rounds, and marks whether the candidate has ever been ranked or ranked first by \(P\). For each candidate-level evidence item, the memory updates phenotype support, phenotype contradiction, and evidence notes. If \(P\) marks a candidate as supported and no strong phenotype contradiction is present, the candidate can remain or become \texttt{active}. If the phenotype evidence strongly contradicts a candidate, the candidate is marked as requiring validation rather than immediately hard-suppressed.

\textbf{Mechanism updates.}
For each mechanism assessment, the memory records the mechanism verdict, verdict strength, suppression recommendation, and mechanism-side evidence notes. A supported candidate receives a mechanism-support flag. A rejected candidate receives contradiction evidence, but mechanism rejection alone does not remove a phenotype-visible candidate. New single-label mechanism proposals are inserted as \texttt{needs\_phenotype\_support}. Family-level proposals create or update entries for the candidate labels in the family subset and attach the corresponding family name.

\textbf{Promotion and suppression.} The update rules distinguish soft disagreement from hard suppression. A candidate can be promoted from \texttt{needs\_phenotype\_support} to \texttt{active} when it receives phenotype support or when mechanism support is consistent with non-contradictory phenotype evidence. Conversely, hard suppression requires both strong phenotype contradiction and strong mechanism rejection. In simplified form, hard suppression is applied only when
\[
\begin{aligned}
&\texttt{phenotype\_contradiction}=\texttt{strong},\\
&\texttt{mechanism\_verdict}=\texttt{reject},\\
&\texttt{mechanism\_strength}=\texttt{strong},\\
&\texttt{suppression\_recommendation}=\texttt{suppress}.
\end{aligned}
\]
This rule prevents either agent from unilaterally deleting a plausible
candidate.

\textbf{Use of reliability metadata.} Reliability metadata from the calibration layer can attach soft warnings to a candidate, for example if it is supported mainly by low-trust anchors or belongs to a weak phenotype class. However, reliability metadata alone does not create a candidate, suppress a candidate, or make a candidate final-eligible. It only modulates how P, M, the controller, and the arbiter interpret the candidate's evidence state.

\textbf{Final eligibility.}
After each update, the system recomputes whether each candidate is visible to the controller and eligible for final prediction. A final-eligible candidate must be canonical, not hard-suppressed, and supported by adequate evidence in the memory. Candidates introduced only by mechanism-side family reasoning, noncanonical labels, and candidates supported only by calibration metadata are not final-eligible. If no candidate remains adequately supported, the final gate may output \texttt{novel} rather than forcing selection from a weak known-MOA candidate.

These update rules are intentionally asymmetric. Phenotype evidence initializes and grounds candidates, mechanism evidence critiques and proposes alternatives, and reliability metadata contextualizes evidence quality. The shared memory therefore acts as the interface through which heterogeneous evidence sources are integrated without collapsing them into a single retrieval score or a free-form agent vote.

\subsection{Controller-Guided Inference Refinement}
\label{app:controller_guided_refinement}
PhenoAIR uses a controller to decide what evidence should be collected next and an arbiter to convert the final candidate memory into a prediction. The controller is deliberately separated from the agents: the phenotype and mechanism agents produce structured evidence, while the controller only selects the next refinement operation based on the current memory state and trajectory. This section describes the controller action space, the controller decision logic, the arbiter logic, and the offline self-evolving procedure used for open-set inference.

\subsubsection{Controller Action Space}
\label{app:controller_action_space}
At refinement step \(t\), the controller selects an action
$a^{(t)} \in \mathcal{A}$ based on the current candidate memory \(\mathcal{H}_q^{(t)}\), reliability context \(r_q\), and trajectory summary \(\tau_q^{(t)}\). The action space contains a small set of evidence-collection and termination operations. We group them into four macro families.
\textbf{Phenotype expansion.}
These actions request additional phenotype-side evidence when the current
retrieval neighborhood is ambiguous, low-confidence, or insufficiently broad:
\[
\texttt{EXPAND\_K}, \qquad
\texttt{GET\_BROADER\_SUMMARY}.
\]
The former increases the retrieval depth, while the latter asks the phenotype agent to summarize broader MOA distributions beyond the current local view.

\textbf{Candidate validation.} These actions ask the phenotype agent to validate specific hypotheses already present in the candidate memory:
\[
\texttt{CHECK\_CANDIDATE\_IN\_NEIGHBORHOOD}, \qquad
\texttt{CHECK\_ACTIVE\_CANDIDATE\_SET}.
\]

They are used when a candidate is visible but its phenotype grounding is insufficient, or when several active candidates need to be compared under the same retrieval context.

\textbf{Group-level validation and mechanism reframing.}
When mechanism-side evidence points to a family rather than a single canonical MOA, the controller can request group-level phenotype inspection:
\[
\texttt{CHECK\_GROUP\_IN\_NEIGHBORHOOD}.
\]
It can also request that the mechanism agent return a family-level assessment:
\[
\texttt{REQUEST\_GROUP\_MODE\_FROM\_M}.
\]

These actions are useful when the system needs to distinguish fine-grained label confusion from broader mechanistic support.

\textbf{Reference inspection and termination.}
The controller may inspect a retrieved reference anchor when a candidate depends heavily on one or a few potentially risky anchors:
$\texttt{INSPECT\_REFERENCE}.$

Finally, it may terminate refinement:
$\texttt{STOP}.$
Termination does not itself choose the final prediction; it only passes the final memory and trajectory to the arbiter. This action space is intentionally coarse-grained. The controller does not directly edit predictions or candidate labels. It only decides which evidence view should be collected next.

\subsubsection{Controller Decision Logic}
\label{app:controller_decision_logic}
The controller policy is implemented as an interpretable signal-driven function:
\[
a^{(t)}
=
\pi\big(
\mathcal{H}_q^{(t)}, r_q, \tau_q^{(t)}
\big).
\]
Rather than relying on free-form LLM deliberation, the controller operates over a compact set of runtime signals derived from the current phenotype output, mechanism output, candidate memory, and action history. Examples include the reliability of the phenotype neighborhood, the agreement between phenotype and mechanism evidence, whether the top phenotype candidate has been rejected by mechanism evidence, whether candidate validation has already been performed, and whether the trajectory has already explored broader phenotype evidence.

The decision logic follows a simple priority structure. First, terminal guards prevent unproductive refinement, such as repeated action cycles or exhausting the round budget. Second, if a promising but insufficiently grounded candidate or candidate group exists, the controller prioritizes validation. Third, if the phenotype evidence is weak, unstable, or based on a narrow neighborhood, the controller requests broader phenotype evidence. Fourth, if the mechanism agent suggests that evidence is better interpreted at the family level, the controller requests group-level validation or mechanism-side reframing. If no additional evidence operation is expected to change the memory state, the controller stops.

The controller can be viewed as selecting among evidence views rather than making predictions. A typical decision has the form:
\[
\mathrm{condition}(\mathcal{H}_q^{(t)}, r_q, \tau_q^{(t)})
\Rightarrow
\mathrm{action}(\mathrm{arguments}),
\]
where the condition is defined over exposed runtime signals and the action is one of the allowed operations above. We use this rule-based form for two reasons. First, it makes the inference protocol auditable: each refinement step can be traced to a controller action and a small set of triggering signals. Second, it prevents the controller from directly introducing unsupported labels or overriding agent evidence.

Importantly, the controller is not designed as a large hand-tuned cascade. Its rules express generic refinement behaviors: validate unresolved candidates, expand unreliable phenotype evidence, inspect risky anchors, request family-level reasoning when the mechanism evidence is coarse, and stop when additional evidence is unlikely to alter the candidate memory. These behaviors are applied uniformly across queries.

\subsubsection{Arbiter Logic}
\label{app:arbiter_logic}
After the controller stops at step \(T\), the arbiter produces the final output: \[ \hat{y}_q = \mathrm{Arbiter} \big( \mathcal{H}_q^{(T)}, \tau_q^{(T)} \big). \]

The arbiter is an LLM-based component, but it is not an evidence-gathering agent. It cannot call retrieval, structure, target-prediction, or reference inspection tools. Its input is restricted to the final candidate memory, the trajectory summary, and the structured outputs already produced by the phenotype and mechanism agents. The arbiter follows three principles. First, it selects from candidate states, not from raw retrieved labels. A candidate must be present in the memory and must have an auditable evidence history. Second, it distinguishes support from visibility. A label may appear in the retrieval neighborhood or in mechanism reasoning, but it is not final-eligible unless it receives adequate grounding in the candidate memory. Third, it respects deterministic eligibility checks after the LLM decision. The deterministic gate filters the arbiter output before evaluation. A selected canonical candidate must satisfy the following high-level requirements:
\[
\begin{aligned}
&\text{the candidate is in the task label space } \mathcal{M},\\
&\text{the candidate is not suppressed},\\
&\text{the candidate is supported by phenotype or mechanism evidence},\\
&\text{the candidate does not have unresolved strong contradiction.}
\end{aligned}
\]
If the arbiter selects a candidate that fails these checks, the system falls back to the best eligible candidate when one exists. If no canonical candidate is adequately supported, the final gate returns \texttt{novel}. Thus, \texttt{novel} is not treated as a free-form label produced by the arbiter, but as the outcome of a final eligibility test over known-MOA candidates. This separation is important for open-set MOA inference. The arbiter can reason over the full trajectory, but the final decision is still constrained by candidate memory and deterministic validity checks.

\subsubsection{Self-Evolving Procedure}

\label{app:self_evolving_procedure}
For open-set inference, the main challenge is deciding when weak evidence for known MOAs should be interpreted as insufficient support for any known MOA. To improve this decision boundary, we use an offline self-evolving procedure over a held-out development set. This procedure does not update LLM parameters, does not modify the phenotype or mechanism tools, and does not use the test set. It only updates two bounded artifacts: the controller rule registry and the mutable decision-heuristic section of the arbiter. The self-evolving loop operates over development trajectories generated by the current multi-agent system. Each trajectory contains the query metadata, final prediction, phenotype-agent summaries, mechanism-agent assessments, controller action history, candidate memory states, and final arbiter output. Since full trajectories are long, they are compressed into case packets that retain the key signals needed for diagnosis, such as the final candidate memory, the sequence of controller decisions, top phenotype candidates across rounds, mechanism verdicts, and whether known-MOA candidates remained adequately supported. Each evolution round has two LLM-assisted steps.

\textbf{Step 1: failure pattern analysis.}
The first step diagnoses systematic errors. Given compact trajectory packets from failure cases and a small set of success guardrails, the LLM clusters failures by recurring root cause. The output is a structured JSON summary of failure clusters. Each cluster contains a description, example cases, a distinguishing signature over phenotype behavior, mechanism behavior, controller trajectory, memory state, and final arbitration behavior, and an assessment of which evolvable surface is implicated:
\[\{
\texttt{controller\_rule\_gap},\quad
\texttt{arbiter\_heuristic\_gap},\quad
\texttt{agent\_interaction\_gap},\quad
\texttt{data\_limit\_or\_one\_off}\}.
\]
Clusters that are rare or appear to be one-off data limitations are not treated as primary targets for artifact updates. Novel-case failures are analyzed separately from closed-set failures so that open-set detection errors are not confused with ordinary wrong-label errors.

\textbf{Step 2: artifact proposal.} For selected systematic clusters, a second LLM call proposes bounded artifact updates. The allowed outputs are:
\[\{
\text{formal controller rules}, \quad
\text{arbiter heuristic updates}, \quad
\text{runtime signal extension requests}, \quad
\text{retention decisions}\}.
\]
Controller rules must follow a fixed schema. Each rule has a rule identifier, priority, description, conditions over whitelisted runtime signals, and one allowed controller action. Rules cannot access arbitrary hidden state or directly edit candidate memory. If the LLM identifies a useful trigger that is not exposed as a runtime signal, it must output a signal-extension request rather than an executable rule. Arbiter updates are similarly constrained. The arbiter prompt is divided into immutable sections and one mutable decision-heuristic section. The LLM may propose a replacement for the mutable heuristic section, but cannot change the arbiter role, input format, output schema, or immutable constraints. This keeps self-evolution focused on decision criteria rather than redefining the task.

\textbf{Validation and materialization.} Before an artifact is used, it is validated against schema and semantic constraints. Controller rules must use allowed condition roots, allowed operators, whitelisted runtime signals, and allowed actions with the required arguments. Invalid or non-executable rules are blocked and saved for inspection. Valid artifacts are materialized into the next development run.

\textbf{Cross-round retention.} After the first evolution rounds, the procedure also compares outcomes across rounds. It tracks which cases improved, which regressed, and which changed without affecting correctness. Subsequent LLM analysis can then recommend keeping, disabling, or revising prior artifacts. This cross-round step is used to reduce the chance that a rule which fixes one failure cluster harms existing success cases.

The self-evolving procedure can therefore be summarized as offline, schema-constrained artifact evolution:
\[\text{trajectories}
\rightarrow
\text{failure clusters}
\rightarrow
\text{validated controller/arbiter artifacts}
\rightarrow
\text{development rerun}.
\]
The final test evaluation uses the selected artifacts fixed in advance. This makes the procedure different from test-time reflection: the system does not adapt to individual test examples using their outcomes, and all evolution is performed before final evaluation.


\section{Benchmark and Evaluation Details}
\label{app:benchmark}
We describe how the benchmark was constructed from raw data, how 
ground-truth MOA labels were curated, and how predictions are evaluated across the three settings.
\subsection{Benchmark Construction}
\label{app:dataset}
The morphological profiles in our benchmark are drawn from the  \texttt{cpg16} dataset of the JUMP Cell Painting Consortium \citep{chandrasekaran2023jump}, with profiles aggregated across five independent laboratory sources (\texttt{source\_1}, \texttt{source\_2}, \texttt{source\_3}, \texttt{source\_5}, and  \texttt{source\_7}). MOA annotations are obtained from the Broad Institute Drug Repurposing Hub \citep{corsello2017drug}, which 
provides hand-curated MOA, target, and clinical-phase annotations for over 6{,}000 compounds. Compounds in the Cell Painting dataset  are matched to MOA annotations via JCP2022 identifiers. Well-level profiles undergo MAD-based feature selection following  the standard JUMP processing pipeline\footnote{https://cellpainting-gallery.s3.amazonaws.com/index.html\#cpg0016-jump/}. Profiles are aggregated to  the treatment level via per-source median across replicate wells  and plates, and to the global consensus via cross-source median.  Per-source treatment-level profiles are retained for downstream multi-source analysis. 

Drug Repurposing Hub annotations are multi-label and use a free-text 
format with multiple MOA strings concatenated by ``|'' (e.g., 
``\textit{Bcr-Abl kinase inhibitor|SRC inhibitor|tyrosine kinase 
inhibitor}''). We parse these into atomic labels and apply two 
curation steps to ensure clean evaluation:
\begin{itemize}
    \item {Hierarchical redundancy removal.} A compound annotated 
    with both a specific MOA and its parent category 
    (e.g., \textit{EGFR inhibitor} and \textit{tyrosine kinase inhibitor}) 
    introduces label-set redundancy that artificially inflates 
    multi-label evaluation. We identify and remove compounds whose 
    label sets contain such hierarchical subsumption relationships.
    
    \item {Representative MOA designation.} For Top-1 evaluation 
    and retrieval-quality analyses, we designate the first-listed 
    label as the representative MOA. Because hierarchical redundancies 
    are removed in the previous step, the remaining multi-label 
    annotations are independent and non-hierarchical, making the 
    representative-label choice inconsequential to the primary 
    evaluation metrics.
\end{itemize}

We construct three settings:
\begin{itemize}
    \item \textbf{MoA-Verified.} The reference set includes compounds profiled across multiple experimental sources, providing more reliable and reproducible phenotype evidence. MOA classes containing fewer than two distinct reference compounds are excluded to reduce instability from singleton classes and ensure that retrieval reflects reproducible mechanistic structure rather than isolated compounds.

    \item \textbf{MoA-Extended.} The reference set pools all available compounds regardless of source count, substantially expanding MOA coverage at the cost of increased heterogeneity and noisier phenotype evidence. As in MoA-Verified, only MOA classes with at least two distinct reference compounds are retained.

    \item \textbf{MoA-Novel.} An open-set evaluation setting constructed using the same reference set as MoA-Verified, but with query compounds whose MOA classes are absent from the reference label space. A test compound is labeled novel only if none of its curated MOA labels appears in the reference label space after synonym normalization and hierarchical redundancy removal. This setting evaluates whether the system can recognize when available evidence is insufficient to support any known mechanism and correctly classify the query as novel.
\end{itemize}
We standardize compounds by InChIKey and ensure that no identical compound appears in both reference and test sets. To assess potential analog leakage, we compute ECFP-based Tanimoto similarity between each test compound and its nearest reference compound. We screen for potential near-duplicate or same-active-series overlap by the maximum ECFP Tanimoto similarity from each test compound to the reference set. Test compounds whose nearest reference exceeds a predefined high-similarity threshold are flagged as potential analog-overlap cases. Our curated benchmarks are smaller than the full JUMP compound collection because MOA prediction requires reliable mechanism annotations, non-redundant label curation, and sufficient reference anchors per class. We therefore prioritize reference quality and controlled evaluation over maximizing dataset size. This choice is particularly important for Cell Painting MOA prediction, where noisy or hierarchically redundant labels can inflate apparent performance or obscure whether a method is reasoning over reliable phenotypic evidence. MoA-Extended provides a larger and noisier setting with broader MOA coverage, while MoA-Verified and MoA-Novel are intentionally compact settings for controlled evaluation of cross-source reliability and open-set recognition. We view further scaling to larger MOA ontologies and additional split constructions as an important direction for future work.

\textbf{Leakage controls.} Several controls are applied to prevent label and structure leakage:
\begin{itemize}

    \item \textbf{No compound overlap between reference and test sets.} 
    Reference and test compounds (identified by JCP2022 IDs) are strictly disjoint in all evaluation settings.
    \item \textbf{Novel queries are mechanistically out-of-scope.} For MoA-Novel, we additionally verify that novel query MOAs do not belong to mechanistic families represented in the reference label space, preventing trivial lexical or hierarchical overlap with known classes.
   \item \textbf{Self-evolving uses a held-out development set.} Self-evolving refinement is performed using a fixed development split of 30 compounds that does not overlap with the final test set.

    \item \textbf{External tool outputs are filtered for direct leakage.} Information returned by external tools is carefully screened to remove fields that directly reveal ground-truth MOA annotations or deterministic identifier-based matches, ensuring that predictions rely on inference over retrieved evidence rather than explicit label exposure.
\end{itemize}

\subsection{Evaluation Protocol}
\textbf{Metrics.}
We report two primary metrics on MoA-Verified and MoA-Extended:
\begin{itemize}
    \item \textbf{Top-1 Accuracy.} A prediction is correct if the primary predicted MOA is semantically equivalent to the representative ground-truth MOA associated with the compound.
    \item \textbf{Hit Rate (Any-match).} A prediction is correct if any predicted MOA (primary or secondary) is semantically equivalent to any MOA in the ground-truth label set.
\end{itemize}
For MoA-Novel, we additionally report:
\begin{itemize}
    \item \textbf{Overall Accuracy.} Whether the prediction, including a possible \texttt{novel} output, is correct.
    \item \textbf{Precision, Recall, F1 (Novel).} Novel-detection metrics treating \texttt{novel} prediction as a binary classification task over out-of-scope and in-scope queries.
\end{itemize}

\paragraph{LLM-as-judge for semantic equivalence.}
LLM-based methods may produce free-text MOA descriptions that differ lexically from canonical labels (e.g., ``histone deacetylase inhibitor'' vs.\ ``HDAC inhibitor''). We therefore use an LLM-as-judge to evaluate semantic equivalence between predicted and ground-truth MOA labels for both Top-1 and Any-match evaluation. We validate judge reliability against human annotations on a randomly sampled subset of prediction-ground-truth pairs and observe 98\% agreement.

\subsection{Implementation and Reproducibility}
\label{app:imple}
\paragraph{LLM backbones.}
We evaluate PhenoAIR using GPT-5.1 (Azure OpenAI) and Claude 4 Sonnet (Anthropic). Lightweight inference configurations are used for evaluation-only utilities such as semantic judging, while agentic components use reasoning-oriented configurations for multi-step tool-based inference.

\paragraph{Cheminformatics tools.}
We use \textbf{RDKit} for chemical structure parsing, fingerprint computation (Morgan fingerprints with radius 2 and 1024 bits), Tanimoto similarity, and pharmacophore-pattern detection. SMILES strings are canonicalized with RDKit prior to all structural operations.
\paragraph{External knowledge resources.}
The Mechanism Agent and offline calibration pipeline use several public bioinformatics resources under a leakage-controlled access protocol. For reference compounds, we use external annotations to characterize retrieved anchors and support offline reliability calibration. For query/test compounds, however, we do not expose known targets, known MOA annotations, pharmacological descriptions, or database entries that could directly reveal the ground-truth mechanism. Query-side access is restricted to structure-derived inputs, including SMILES/InChIKey identifiers, basic physicochemical descriptors, structural similarity, molecular fingerprints, pharmacophore matches, and controlled in-silico target predictions.
The resources are used as follows:
\begin{itemize}
    \item \textbf{ChEMBL}\footnote{https://www.ebi.ac.uk/chembl/}: bioactivity data and structurally similar analogs of query compounds.
    \item \textbf{DrugBank} \footnote{https://go.drugbank.com/}: known drug targets and pharmacological annotations.
    \item \textbf{PubChem} \footnote{https://pubchem.ncbi.nlm.nih.gov/}: compound metadata and cross-database identifier resolution.
    \item \textbf{SwissTargetPrediction} \footnote{https://www.swisstargetprediction.ch/}: in-silico target prediction based on structural similarity to known bioactive compounds.
\end{itemize}

\paragraph{Cell Painting features.}
We evaluate multiple morphological feature backbones, including CellProfiler \cite{carpenter2006cellprofiler}, CA-MAE~\cite{kraus2024masked}, and CellCLIP \cite{lucellclip}. For CA-MAE and CellCLIP, we use ViT-Base encoders pretrained on raw Cell Painting images from the five selected sources, covering approximately 3.3 million images. Each model is pretrained for 100 epochs on H100 GPUs, requiring approximately 120 GPU hours. Profile preprocessing and aggregation pipelines are applied consistently across feature representations unless otherwise specified.

\paragraph{Inference budget.}
The maximum number of refinement rounds is capped at $T_{\max}=8$. Self-evolving refinement terminates when no further rule updates are produced or when development-set performance no longer improves.  Runtime statistics and API usage summaries are provided in Appendix~\ref{app:cost}.

\section{Prompt Templates}
\label{app:prompts}

This section provides representative prompt templates used in PhenoAIR. The prompts are designed to enforce role-specific reasoning constraints, separate phenotype and mechanism evidence, and support iterative refinement through controller-guided interaction. For brevity, low-level orchestration details and implementation-specific tool-call formatting are omitted.
\begin{tcolorbox}[
fonttitle=\bfseries\small,
  title={Phenotype Agent Prompt},
  boxrule=0.5pt, arc=2pt,
  left=6pt, right=6pt, top=3pt, bottom=3pt,breakable
]
{\small

You are the Phenotype Agent (P) in a multi-agent system for drug mechanism of action (MOA) prediction.

Your role is to interpret phenotype-side Cell Painting retrieval evidence. Phenotype retrieval results include calibrated reliability information in addition to raw similarity scores. Retrieved reference compounds should be treated as phenotype anchors for interpreting the local neighborhood of the query compound, rather than as direct labels.

You are allowed to:
\begin{itemize}
    \item inspect nearest-neighbor retrieval results,
    \item inspect broader MOA distributions,
    \item verify whether specific MOAs are supported by the phenotype neighborhood,
    \item inspect retrieved reference compounds for phenotype interpretation.
\end{itemize}

Your job is to:
\begin{itemize}
    \item propose ranked MOA candidates based on phenotype evidence,
    \item assess retrieval reliability and neighborhood consistency,
    \item identify ambiguity, instability, or potential confounding patterns,
    \item revise candidate rankings when additional phenotype evidence becomes available.
\end{itemize}

Important constraints:
\begin{itemize}

    \item Do not use structure-based evidence, pharmacophore evidence, or in-silico target predictions as independent decision criteria.

    \item Do not assume that the top-ranked neighbor directly determines the answer.

    \item Predictions must remain grounded in phenotype-side retrieval evidence and the structure of the local neighborhood.

    \item Retrieved reference compounds may be used to interpret phenotype patterns, but should not act as standalone mechanistic proof.

    \item When phenotype evidence is weak or mixed, maintain a broader candidate set rather than collapsing to a single hypothesis prematurely.

    \item When the controller provides a constrained candidate set, perform phenotype-side discrimination only within that set.
\end{itemize}
Candidate memory discipline:

You will receive a memory summary containing previously proposed candidates, supporting evidence, and suppression signals from prior rounds. The memory supports iterative refinement and consistency tracking, but should not be treated as a final decision oracle.
Label discipline:
\begin{itemize}

    \item Only output labels from the predefined benchmark MOA label space.

    \item Do not invent new labels, paraphrase canonical labels, or produce free-form mechanistic descriptions.

    \item Do not output \texttt{novel moa}; novelty decisions are reserved for the final Arbiter.

    \item When evidence is insufficient, lower confidence and retain uncertainty rather than generating unsupported labels.
\end{itemize}

\normalsize
}
\end{tcolorbox}

\begin{tcolorbox}[
fonttitle=\bfseries\small,
  title={Mechanism Agent Prompt},
  boxrule=0.5pt, arc=2pt,
  left=6pt, right=6pt, top=3pt, bottom=3pt,breakable
]
{\small

You are the Mechanism Agent (M) in a multi-agent system for drug mechanism of action (MOA) prediction.

Your role is to evaluate phenotype-derived MOA candidates using structure-driven mechanistic evidence.

You are allowed to:

\begin{itemize}

    \item inspect compound structure and physicochemical properties,

    \item inspect pharmacophore patterns,

    \item inspect in-silico target predictions,

    \item inspect structurally similar compounds and their mechanistic annotations,

    \item inspect known reference compounds and associated mechanism information.

\end{itemize}

You are instructed to:

\begin{itemize}

    \item evaluate MOA candidates proposed by the Phenotype Agent,

    \item determine whether mechanism-side evidence supports, rejects, or remains uncertain about each candidate,

    \item summarize overall agreement with phenotype-side reasoning,

    \item determine the appropriate granularity of mechanistic conclusions.

\end{itemize}

Mechanistic conclusions may operate at different levels of specificity:

\begin{itemize}

    \item If the evidence strongly supports a specific canonical MOA label, return a single-candidate conclusion.

    \item If the evidence only supports a mechanistic family or a small subset of related labels, return a constrained candidate group rather than forcing a fine-grained label decision.

\end{itemize}

Important constraints:

\begin{itemize}

    \item Do not inspect raw Cell Painting retrieval results or phenotype neighborhoods directly.

    \item Do not output a full ranking over all MOA labels.

    \item Do not act as the final decision-maker.

    \item Treat mechanistic evidence as complementary evidence rather than a standalone oracle overriding phenotype observations.

    \item If mechanism-side evidence strongly supports an allowed MOA absent from the current phenotype candidate list, you may still propose it for phenotype-side validation.

    \item Very weak structural similarity alone should not override stronger target-level or pharmacological evidence.

\end{itemize}

You receive a memory summary containing previously proposed candidates, supporting evidence, and suppression signals from prior rounds. The memory is intended to support iterative refinement and consistency tracking rather than directly determine final predictions.

Label discipline:

\begin{itemize}

    \item Only output labels from the predefined benchmark MOA label space.

    \item Do not invent new labels, paraphrase canonical labels, or generate free-form mechanistic descriptions as candidate labels.

    \item If evidence supports only a broader mechanistic family, map the evidence to the corresponding canonical benchmark labels rather than introducing new family names.

    \item Do not output \texttt{novel moa}; novelty decisions are reserved for the final Arbiter.

\end{itemize}

\normalsize
}
\end{tcolorbox}
\begin{tcolorbox}[
fonttitle=\bfseries\small,
  title={Arbiter Agent Prompt},
  boxrule=0.5pt, arc=2pt,
  left=6pt, right=6pt, top=3pt, bottom=3pt,breakable
]
{\small
You are the Arbiter Agent (A) in a multi-agent system for drug mechanism of action (MOA) prediction.

Your role is to make the final decision based on structured outputs from the Phenotype Agent, the Mechanism Agent, candidate memory, and controller history. Your main value is meta-level confidence calibration and generating an interpretable final rationale, rather than redoing the phenotype or mechanism analysis from scratch.

You are instructed to:

\begin{itemize}

    \item determine the final predicted MOA,

    \item calibrate confidence based on the full refinement trajectory,

    \item include alternative candidates when evidence remains ambiguous,

    \item output \texttt{novel moa} when no task-allowed canonical label is adequately supported,

    \item provide a concise rationale that downstream researchers can audit.

\end{itemize}

Confidence calibration:

\begin{itemize}

    \item Assign high confidence when phenotype retrieval is reliable, mechanism evidence supports the selected candidate, and the candidate is stable across refinement rounds.

    \item Assign medium confidence when evidence improves after refinement but remains partially ambiguous, when mechanism evidence is uncertain, or when mechanism evidence supports only a candidate group rather than a specific label.

    \item Assign low confidence when candidate rankings oscillate, retrieval reliability remains limited, mechanism and phenotype evidence remain in conflict, or the final choice relies on marginal support.

\end{itemize}

Mechanistic resolution:

\begin{itemize}

    \item If the Mechanism Agent supports a specific canonical MOA, treat this as strong evidence only when it is not contradicted by phenotype evidence.

    \item If the Mechanism Agent supports only a candidate group or mechanistic family, treat this as weaker evidence and do not upgrade confidence based on mechanism evidence alone.

    \item If mechanism evidence suggests a non-canonical or overly fine-grained label, use it only as supporting evidence for the nearest canonical candidate already present in memory.

\end{itemize}

Final-choice rules:

\begin{itemize}

    \item Candidate memory records which hypotheses appeared during refinement and which were supported, challenged, or suppressed. It should not be treated as a ranking by itself.

    \item Do not choose solely by phenotype score when the phenotype trajectory is weak, sparse, or non-specific.

    \item Do not choose solely by mechanism support when a phenotype-supported candidate is stable and not specifically rejected.

    \item Prefer a mechanism-supported candidate over a phenotype-favored candidate only when mechanism evidence is specific, the candidate is canonical, and phenotype evidence does not strongly contradict it.

    \item For same-family candidate labels, prefer the label most specifically supported by mechanism evidence only when phenotype evidence cannot distinguish among them.

    \item Output exactly \texttt{novel moa} when the integrated evidence indicates that no task-allowed canonical candidate is sufficiently supported.

\end{itemize}

Important constraints:

\begin{itemize}

    \item Do not reinterpret the task from scratch.

    \item Do not use external tools.

    \item Do not invent candidates that were not introduced by the Phenotype Agent, Mechanism Agent, controller, or candidate memory.

    \item The final \texttt{primary\_moa} must be either a task-allowed canonical label or exactly \texttt{novel moa}.

    \item Do not output \texttt{ABSTAIN}, \texttt{no\_supported\_moa}, or an empty string as the final label.

    \item Do not select archived or hard-suppressed candidates unless no viable canonical alternative remains.

\end{itemize}

Before submitting the final prediction, perform a brief decision audit:

\begin{itemize}
    \item Is the selected label canonical and not hard-suppressed?
    \item Was it supported by phenotype evidence, mechanism evidence, or both?
    \item Was it strongly rejected by either agent, and if so, why is it still selected?
    \item Are stronger alternatives suppressed or unsupported?
    \item Is the evidence resolved at the canonical-label level, or only at the family level?

\end{itemize}

Use this audit only for calibration and final choice. Do not redo the Phenotype Agent or Mechanism Agent analysis from scratch.
\normalsize
}
\end{tcolorbox}
\newpage
\begin{algorithm}[H]
\caption{PhenoAIR Inference with Global Reliability Calibration}
\label{alg:phenoair_inference}
\begin{algorithmic}[1]
\Require Reference set $\mathcal{R}$ with Cell Painting profiles and MOA labels;
query compound $q$ with Cell Painting profile and structure;
canonical MOA label space $\mathcal{M}$;
maximum refinement rounds $T_{\max}$
\Ensure Final prediction $\hat{y}_q \in \mathcal{M} \cup \{\texttt{novel}\}$

\Statex \textbf{Global pre-computation before query-time inference}
\State Build source-specific reference profiles from well-level features.
\State Construct reference reliability memory $\mathcal{B}$:
\Statex  use offline LLM-assisted code execution to analyze source-level,
MOA-level, reference-anchor-level, and MOA-pair reliability.
\State Construct mechanistic family ontology $\mathcal{G}$:
\Statex  group canonical MOA labels into coarse mechanistic families using
external biomedical evidence.
\State Define a fixed calibrated retrieval rule $\mathrm{Calib}(\cdot)$ using the reference reliability memory $\mathcal{B}$.
\Statex \Comment{The above steps are global and are not repeated for each test query.}

\Statex \textbf{Query-time inference for query $q$}
\State Retrieve raw phenotype neighborhood $\mathcal{N}(q)$ from the reference set.
\State Compute calibrated neighborhood
\[
\tilde{\mathcal{N}}(q) \leftarrow
\mathrm{Calib}(\mathcal{N}(q), \mathcal{B}).
\]
\State Build query-level reliability context $r_q$ from calibrated retrieval and
reference memory.
\State Initialize candidate memory $\mathcal{H}_q^{(0)} \leftarrow \emptyset$.
\State Initialize trajectory log $\tau_q^{(0)} \leftarrow \emptyset$.

\For{$t = 0,1,\ldots,T_{\max}$}
    \State Run phenotype agent:
    \[
    P_q^{(t)} \leftarrow
    P(\tilde{\mathcal{N}}(q), r_q, \mathcal{H}_q^{(t)}, \tau_q^{(t)}).
    \]
    \State Update candidate memory with phenotype evidence:
    \[
    \mathcal{H}_q^{(t+\frac{1}{3})}
    \leftarrow
    \mathrm{UpdateP}(\mathcal{H}_q^{(t)}, P_q^{(t)}).
    \]

    \State Run mechanism agent:
    \[
    M_q^{(t)} \leftarrow
    M(q, \mathcal{G}, \mathcal{H}_q^{(t+\frac{1}{3})}, P_q^{(t)}).
    \]
    \State Update candidate memory with mechanism evidence:
    \[
    \mathcal{H}_q^{(t+\frac{2}{3})}
    \leftarrow
    \mathrm{UpdateM}(\mathcal{H}_q^{(t+\frac{1}{3})}, M_q^{(t)}).
    \]

    \State Select controller action:
    \[
    a^{(t)}
    \leftarrow
    \pi(\mathcal{H}_q^{(t+\frac{2}{3})}, r_q, \tau_q^{(t)}).
    \]

    \State Append $(P_q^{(t)}, M_q^{(t)}, a^{(t)}, \mathcal{H}_q^{(t+\frac{2}{3})})$
    to trajectory log $\tau_q$.

    \If{$a^{(t)} = \texttt{STOP}$}
        \State $\mathcal{H}_q^{(T)} \leftarrow \mathcal{H}_q^{(t+\frac{2}{3})}$.
        \State \textbf{break}
    \Else
        \State Execute controller action $a^{(t)}$ to obtain additional evidence view.
        \Statex \quad Examples: expand phenotype neighborhood, validate candidate,
        validate candidate group, inspect reference anchor, or request mechanism
        group mode.
        \State Update available evidence context for the next round.
        \State $\mathcal{H}_q^{(t+1)} \leftarrow \mathcal{H}_q^{(t+\frac{2}{3})}$.
    \EndIf
\EndFor

\State Run arbiter over final memory and trajectory:
\[
y_q^{\mathrm{arb}}
\leftarrow
A(\mathcal{H}_q^{(T)}, \tau_q^{(T)}).
\]
\State Apply deterministic final eligibility gate:
\[
\hat{y}_q
\leftarrow
\mathrm{FinalGate}(y_q^{\mathrm{arb}}, \mathcal{H}_q^{(T)}).
\]
\State \Return $\hat{y}_q$.
\end{algorithmic}
\end{algorithm}

\newpage

\section*{NeurIPS Paper Checklist}

\begin{enumerate}

\item {\bf Claims}
    \item[] Question: Do the main claims made in the abstract and introduction accurately reflect the paper's contributions and scope?
    \item[] Answer: \answerYes{} 
    \item[] Justification: The abstract and introduction state the paper's scope as Cell Painting-based MOA prediction and frame the main contributions as task reformulation, the multi-agent framework, and a novel benchmark. The claims are supported by the experimental results and ablation studies reported in the main paper.

\item {\bf Limitations}
    \item[] Question: Does the paper discuss the limitations of the work performed by the authors?
    \item[] Answer: \answerYes{}
    \item[] Justification: Section 4.2 discusses the limitations of the proposed approach through an analysis of model failure modes.

\item {\bf Theory assumptions and proofs}
    \item[] Question: For each theoretical result, does the paper provide the full set of assumptions and a complete (and correct) proof?
    \item[] Answer: \answerNA{} 
    \item[] Justification: The paper does not include theoretical results or formal proofs.

    \item {\bf Experimental result reproducibility}
    \item[] Question: Does the paper fully disclose all the information needed to reproduce the main experimental results of the paper to the extent that it affects the main claims and/or conclusions of the paper (regardless of whether the code and data are provided or not)?
    \item[] Answer: \answerYes{} 
    \item[] Justification: Sections 3 and 4 describe the benchmark construction, baselines, evaluation metrics, implementation details, and ablation settings needed to reproduce the main experimental results.

\item {\bf Open access to data and code}
    \item[] Question: Does the paper provide open access to the data and code, with sufficient instructions to faithfully reproduce the main experimental results, as described in supplemental material?
    \item[] Answer: \answerNo{} 
    \item[] Justification: The experiments are based on publicly available Cell Painting profiles and curated MOA annotations, and the paper describes the benchmark construction and evaluation protocol. We do not provide a full code release at submission time.

\item {\bf Experimental setting/details}
    \item[] Question: Does the paper specify all the training and test details (e.g., data splits, hyperparameters, how they were chosen, type of optimizer) necessary to understand the results?
    \item[] Answer: \answerYes{} 
    \item[] Justification: Section 4 describes the experimental settings, including benchmark splits, baselines, evaluation metrics, LLM configurations, and ablation settings needed to interpret the results.

\item {\bf Experiment statistical significance}
    \item[] Question: Does the paper report error bars suitably and correctly defined or other appropriate information about the statistical significance of the experiments?
    \item[] Answer: \answerYes{} 
    \item[] Justification: The main results report averages over multiple runs where stochastic LLM-based methods are used, while deterministic retrieval baselines are reported once; the evaluation protocol and sources of variability are described in Section 4.

\item {\bf Experiments compute resources}
    \item[] Question: For each experiment, does the paper provide sufficient information on the computer resources (type of compute workers, memory, time of execution) needed to reproduce the experiments?
    \item[] Answer: \answerYes{} 
    \item[] Justification: Section 4 and Appendix reports the compute environment and approximate runtime used for the experiments, including LLM inference settings and retrieval/evaluation workloads.
    
\item {\bf Code of ethics}
    \item[] Question: Does the research conducted in the paper conform, in every respect, with the NeurIPS Code of Ethics \url{https://neurips.cc/public/EthicsGuidelines}?
    \item[] Answer: \answerYes{} 
    \item[] Justification: The research uses publicly available biological profiling data and curated annotations, does not involve human subjects or private personal data, and conforms to the NeurIPS Code of Ethics.

\item {\bf Broader impacts}
    \item[] Question: Does the paper discuss both potential positive societal impacts and negative societal impacts of the work performed?
    \item[] Answer: \answerYes{} 
    \item[] Justification: The paper discusses broader impacts, including potential benefits for drug mechanism discovery and risks from over-interpreting model-generated mechanistic hypotheses without experimental validation.
    
\item {\bf Safeguards}
    \item[] Question: Does the paper describe safeguards that have been put in place for responsible release of data or models that have a high risk for misuse (e.g., pre-trained language models, image generators, or scraped datasets)?
    \item[] Answer: \answerNA{} 
    \item[] Justification: The paper does not release high-risk models, scraped datasets, or other assets requiring special misuse safeguards.

\item {\bf Licenses for existing assets}
    \item[] Question: Are the creators or original owners of assets (e.g., code, data, models), used in the paper, properly credited and are the license and terms of use explicitly mentioned and properly respected?
    \item[] Answer: \answerYes{} 
    \item[] Justification: The paper uses publicly available datasets and existing tools/models, and credits the original sources with citations and usage descriptions in the data and implementation sections.

\item {\bf New assets}
    \item[] Question: Are new assets introduced in the paper well documented and is the documentation provided alongside the assets?
    \item[] Answer: \answerYes{} 
    \item[] Justification: The paper introduces curated benchmark splits derived from public datasets, and documents their construction, evaluation protocol, and intended use in the benchmark section.

\item {\bf Crowdsourcing and research with human subjects}
    \item[] Question: For crowdsourcing experiments and research with human subjects, does the paper include the full text of instructions given to participants and screenshots, if applicable, as well as details about compensation (if any)? 
    \item[] Answer: \answerNA{} 
    \item[] Justification: The paper does not involve crowdsourcing experiments or research with human subjects.

\item {\bf Institutional review board (IRB) approvals or equivalent for research with human subjects}
    \item[] Question: Does the paper describe potential risks incurred by study participants, whether such risks were disclosed to the subjects, and whether Institutional Review Board (IRB) approvals (or an equivalent approval/review based on the requirements of your country or institution) were obtained?
    \item[] Answer: \answerNA{} 
    \item[] Justification: The paper does not involve human subjects.

\item {\bf Declaration of LLM usage}
    \item[] Question: Does the paper describe the usage of LLMs if it is an important, original, or non-standard component of the core methods in this research? Note that if the LLM is used only for writing, editing, or formatting purposes and does \emph{not} impact the core methodology, scientific rigor, or originality of the research, declaration is not required.
    \item[] Answer: \answerYes{} 
    \item[] Justification: LLMs are a core component of the proposed method, and the paper describes their roles in phenotype analysis, mechanism evaluation, and final arbitration, as well as the model configurations and prompting/evaluation protocol.

\end{enumerate}

\end{document}